\documentclass{article}
\usepackage[accepted]{icml2020}

\usepackage{times}

\usepackage{soul}
\usepackage{url}
\usepackage[hidelinks]{hyperref}
\usepackage[utf8]{inputenc}
\usepackage[small]{caption}
\usepackage{graphicx}
\usepackage{amsmath}
\usepackage{booktabs}
\usepackage{dsfont}

\usepackage[utf8]{inputenc}
\usepackage{natbib}
\usepackage{amsmath}
\usepackage{amsfonts}
\usepackage{amssymb}
\usepackage{bbold}
\usepackage{bm}
\usepackage{bbm}
\usepackage{color}
\usepackage{subcaption}
\usepackage{graphicx}
\usepackage[sectionbib]{chapterbib}
\usepackage[acronym,smallcaps,nowarn,section,nogroupskip,nonumberlist]{glossaries}
\glsdisablehyper{}

\usepackage[utf8]{inputenc} %
\usepackage[T1]{fontenc}    %
\usepackage{hyperref}       %
\usepackage{url}            %
\usepackage{amsfonts}       %
\usepackage{nicefrac}       %
\usepackage{microtype}      %
\usepackage{amsmath}
\usepackage{graphicx}
\usepackage{subcaption}

\usepackage{wrapfig,lipsum,booktabs}

\usepackage{tikz}
\usetikzlibrary{bayesnet}
\usetikzlibrary{decorations.pathreplacing}

\usepackage{verbatim}

\usepackage{adjustbox}

\usepackage[normalem]{ulem}

\newcommand{\github}{
\href{https://github.com/ecreager/causal-dyna-fair}{
  \texttt{github.com/ecreager/causal-dyna-fair}
  }
}
\newcommand{\MaxProf}{\textsc{MaxProf}}
\newcommand{\DP}{\textsc{DemPar}}
\newcommand{\EqOpp}{\textsc{EqOpp}}

\newcommand{\E}{\mathbb{E}}

\newcommand{\ind}[1]{\mathbb{1} (#1)}

\makeatletter

\pgfdeclareshape{rectangle with diagonal fill}
{
    \inheritsavedanchors[from=rectangle]
    \inheritanchorborder[from=rectangle]
    \inheritanchor[from=rectangle]{north}
    \inheritanchor[from=rectangle]{north west}
    \inheritanchor[from=rectangle]{north east}
    \inheritanchor[from=rectangle]{center}
    \inheritanchor[from=rectangle]{west}
    \inheritanchor[from=rectangle]{east}
    \inheritanchor[from=rectangle]{mid}
    \inheritanchor[from=rectangle]{mid west}
    \inheritanchor[from=rectangle]{mid east}
    \inheritanchor[from=rectangle]{base}
    \inheritanchor[from=rectangle]{base west}
    \inheritanchor[from=rectangle]{base east}
    \inheritanchor[from=rectangle]{south}
    \inheritanchor[from=rectangle]{south west}
    \inheritanchor[from=rectangle]{south east}

    \inheritbackgroundpath[from=rectangle]
    \inheritbeforebackgroundpath[from=rectangle]
    \inheritbehindforegroundpath[from=rectangle]
    \inheritforegroundpath[from=rectangle]
    \inheritbeforeforegroundpath[from=rectangle]

    \behindbackgroundpath{%
        \pgfextractx{\pgf@xa}{\southwest}%
        \pgfextracty{\pgf@ya}{\southwest}%
        \pgfextractx{\pgf@xb}{\northeast}%
        \pgfextracty{\pgf@yb}{\northeast}%
        \ifpgf@diagonal@lefttoright
            \def\pgf@diagonal@point@a{\pgfpoint{\pgf@xa}{\pgf@yb}}%
            \def\pgf@diagonal@point@b{\pgfpoint{\pgf@xb}{\pgf@ya}}%
        \else
            \def\pgf@diagonal@point@a{\southwest}%
            \def\pgf@diagonal@point@b{\northeast}%
        \fi
        \pgfpathmoveto{\pgf@diagonal@point@a}%
        \pgfpathlineto{\northeast}%
        \pgfpathlineto{\pgfpoint{\pgf@xb}{\pgf@ya}}%
        \pgfpathclose
        \ifpgf@diagonal@lefttoright
            \color{\pgf@diagonal@top@color}%
        \else
            \color{\pgf@diagonal@bottom@color}%
        \fi
        \pgfusepath{fill}%
        \pgfpathmoveto{\pgfpoint{\pgf@xa}{\pgf@yb}}%
        \pgfpathlineto{\southwest}%
        \pgfpathlineto{\pgf@diagonal@point@b}%
        \pgfpathclose
        \ifpgf@diagonal@lefttoright
            \color{\pgf@diagonal@bottom@color}%
        \else
            \color{\pgf@diagonal@top@color}%
        \fi
        \pgfusepath{fill}%
    }
}

\newif\ifpgf@diagonal@lefttoright
\def\pgf@diagonal@top@color{white}
\def\pgf@diagonal@bottom@color{gray!30}

\def\pgfsetdiagonaltopcolor#1{\def\pgf@diagonal@top@color{#1}}%
\def\pgfsetdiagonalbottomcolor#1{\def\pgf@diagonal@bottom@color{#1}}%
\def\pgfsetdiagonallefttoright{\pgf@diagonal@lefttorighttrue}%
\def\pgfsetdiagonalrighttoleft{\pgf@diagonal@lefttorightfalse}%

\tikzoption{diagonal top color}{\pgfsetdiagonaltopcolor{#1}}
\tikzoption{diagonal bottom color}{\pgfsetdiagonalbottomcolor{#1}}
\tikzoption{diagonal from left to right}[]{\pgfsetdiagonallefttoright}
\tikzoption{diagonal from right to left}[]{\pgfsetdiagonalrighttoleft}
\makeatother

\begin{document}

\twocolumn[
\icmltitle{Causal Modeling for Fairness in Dynamical Systems}



\icmlsetsymbol{equal}{*}

\begin{icmlauthorlist}
\icmlauthor{Elliot Creager}{to,ve}
\icmlauthor{David Madras}{to,ve}
\icmlauthor{Toniann Pitassi}{to,ve}
\icmlauthor{Richard Zemel}{to,ve}
\end{icmlauthorlist}

\icmlaffiliation{to}{University of Toronto}
\icmlaffiliation{ve}{Vector Institute}

\icmlcorrespondingauthor{Elliot Creager}{creager@cs.toronto.edu}

\icmlkeywords{Machine Learning, Causal Inference, Trustworthy Machine Learning, Feedback Loops, Long-term Fairness, Dynamical Systems, ICML}

\vskip 0.3in
]



\printAffiliationsAndNotice{}  


\begin{abstract}
In many application areas---lending, education, and online recommenders, for example---fairness and equity concerns emerge when a machine learning system interacts with a dynamically changing environment to produce both immediate and long-term effects for individuals and demographic groups. We discuss causal directed acyclic graphs (DAGs) as a unifying framework for the recent literature on fairness in such dynamical systems. We show that this formulation affords several new directions of inquiry to the modeler, where causal assumptions can be expressed and manipulated. We emphasize the importance of computing interventional quantities in the dynamical fairness setting, and show how causal assumptions enable simulation (when environment dynamics are known) and off-policy estimation (when dynamics are unknown) of intervention on short- and long-term outcomes, at both the group and individual levels.
\end{abstract}

\section{Introduction}
How do we design fair policies for complex, evolving systems?
Recently, the literature on fairness in dynamical systems 
has begun exploring the role of algorithmic systems in shaping their environments over time
\citep{hashimoto2018fairness,lum2016predict,ensign2018runaway}.
The key insight from these papers is that the repeated application of
algorithmic tools in a changing environment can have fairness implications in the long-term distinct from those in the short-term.

However, the methods in this literature are quite disparate, with little overlap existing between various works in terms of modeling choices, goals, or assumptions.
This lack of formal similarity is surprising, given that these papers are usually structurally alike: each proposes a dynamics model for a particular domain
(e.g. lending \citep{mouzannar2019fair}, hiring \citep{hu2018short},
recommendations \citep{bountouridis2019siren}), exposes unfairness that arises
from long-term usage of some baseline policy, and then proposes a ``fair''
policy to mitigate some of these biases.

In this paper, we propose unifying the literature on fairness in dynamical
systems via causal directed acyclic graphs (DAGs)
\citep{pearl2009causal,richardson2013single}.
While causal DAGs have been used to study one-shot fair decision-making
\citep{kusner2017counterfactual,kusner2019making, kilbertus2017avoiding}, they
are uncommon in fairness settings involving \emph{sequential} decisions.
We show that several intuitive models of long-term unfairness are naturally expressed using causal DAGs.
We also show that causal reasoning is useful for analyzing models and evaluating policies for these problems.

Our contributions are:
\begin{itemize}
    \item We show that causal DAGs are a unifying framework for the literature
      on fairness in dynamical systems,
      reformulating examples from the literature using \textit{structural causal models} and \textit{policy interventions}.
    \item We demonstrate empirically that when environment dynamics are unknown, causal reasoning can help utilize observational data to improve off-policy estimation and learning.
    \item We show that if dynamics are known, causal DAGs serve as flexible simulators for analyzing policies and models, through extending and investigating model assumptions.
\end{itemize}{}

We proceed as follows.
In Section~\ref{sec:background}, we introduce key background concepts of structural causal models and policy interventions.
In Section \ref{sec:motivating-dags}, we demonstrate the application of causal DAGs to several key concepts in the fairness in dynamical systems literature.
In Section \ref{sec:related-work} we discuss related work in fairness and causality.
In Section \ref{sec:off-policy-eval} we empirically demonstrate that causal modelling can improve off-policy estimation and selection in a dynamical fairness problem, and 
in Section \ref{sec:individual-dynamics} we show how the explication of underlying causal assumptions enables model extension and analysis.\footnote{
Code at \github
}

\section{Background} \label{sec:background}

\subsection{Structual Causal Models}
\newcommand{\mywidth}{0.22\textwidth}
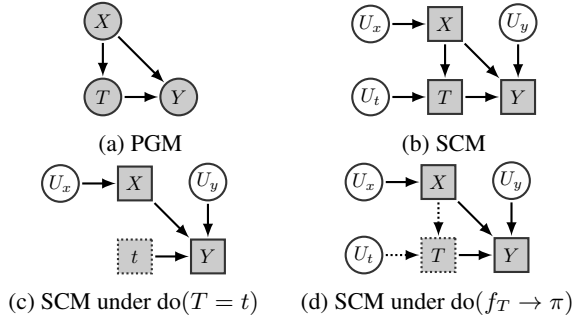
\begin{figure}[t!]
  \centering
  \begin{subfigure}[b]{\mywidth}
    \begin{center}
      \begin{tikzpicture}[]
  \tikzstyle{lat-exo}=[circle, inner sep=1pt, minimum size = 6.5mm, thick, draw=black!80, node distance = 20mm, scale=0.75]
  \tikzstyle{obs-exo}=[circle, fill=black!20, inner sep=1pt, minimum size = 6.5mm, thick, draw=black!80, node distance = 20mm, scale=0.75]
  \tikzstyle{lat-box}=[rectangle, inner sep=1pt, minimum size = 6mm, draw=black!80, thick, node distance = 20mm, scale=0.75]
  \tikzstyle{obs-box}=[rectangle, fill=black!20, inner sep=1pt, minimum size = 6mm, thick, draw=black!80, node distance = 20mm, scale=0.75]
\tikzstyle{inv-box}=[rectangle, fill=black!20, inner sep=1pt, minimum size = 6mm, thick, draw=black!80,densely dotted, node distance = 20mm, scale=0.75]
  \tikzstyle{directed}=[-latex, thick, shorten >=0.5 pt, shorten <=1 pt]
  \tikzstyle{interven}=[directed, densely dotted]
  \node[obs-exo] (Y) {$Y$};
  \node[obs-exo] (T) [left=0.5cm of Y] {$T$};
  \node[obs-exo] (X) [above=0.5cm of T] {$X$};
  \path
    (X) edge [directed] (Y)
    (X) edge [directed] (T)
    (T) edge [directed] (Y)
    ;
\end{tikzpicture}%
      \caption{PGM}
      \label{fig:pgm-treatment}
    \end{center}
  \end{subfigure}
  ~
  \begin{subfigure}[b]{\mywidth}
    \begin{center}
      \begin{tikzpicture}[]
  \tikzstyle{lat-exo}=[circle, inner sep=1pt, minimum size = 6.5mm, thick, draw=black!80, node distance = 20mm, scale=0.75]
  \tikzstyle{obs-exo}=[circle, fill=black!20, inner sep=1pt, minimum size = 6.5mm, thick, draw=black!80, node distance = 20mm, scale=0.75]
  \tikzstyle{lat-box}=[rectangle, inner sep=1pt, minimum size = 6mm, draw=black!80, thick, node distance = 20mm, scale=0.75]
  \tikzstyle{obs-box}=[rectangle, fill=black!20, inner sep=1pt, minimum size = 6mm, thick, draw=black!80, node distance = 20mm, scale=0.75]
\tikzstyle{inv-box}=[rectangle, fill=black!20, inner sep=1pt, minimum size = 6mm, thick, draw=black!80,densely dotted, node distance = 20mm, scale=0.75]
  \tikzstyle{directed}=[-latex, thick, shorten >=0.5 pt, shorten <=1 pt]
  \tikzstyle{interven}=[directed, densely dotted]
  \node[obs-box] (Y) {$Y$};
  \node[obs-box] (T) [left=0.5cm of Y] {$T$};
  \node[obs-box] (X) [above=0.5cm of T] {$X$};
  \node[lat-exo] (Ut) [left=0.5cm of T] {$U_t$} ;
  \node[lat-exo] (Ux) [left=0.5cm of X] {$U_x$} ;
  \node[lat-exo] (Uy) [above=0.5cm of Y] {$U_y$} ;
  \path
    (X) edge [directed] (Y)
    (X) edge [directed] (T)
    (T) edge [directed] (Y)
    (Uy) edge [directed] (Y)
    (Ux) edge [directed] (X)
    (Ut) edge [directed] (T)
    ;
\end{tikzpicture}%
      \caption{SCM}
      \label{fig:scm-treatment}
    \end{center}
  \end{subfigure}

  \begin{subfigure}[b]{\mywidth}
    \begin{center}
      \begin{tikzpicture}[]
  \tikzstyle{lat-exo}=[circle, inner sep=1pt, minimum size = 6.5mm, thick, draw=black!80, node distance = 20mm, scale=0.75]
  \tikzstyle{obs-exo}=[circle, fill=black!20, inner sep=1pt, minimum size = 6.5mm, thick, draw=black!80, node distance = 20mm, scale=0.75]
  \tikzstyle{lat-box}=[rectangle, inner sep=1pt, minimum size = 6mm, draw=black!80, thick, node distance = 20mm, scale=0.75]
  \tikzstyle{obs-box}=[rectangle, fill=black!20, inner sep=1pt, minimum size = 6mm, thick, draw=black!80, node distance = 20mm, scale=0.75]
\tikzstyle{inv-box}=[rectangle, fill=black!20, inner sep=1pt, minimum size = 6mm, thick, draw=black!80,densely dotted, node distance = 20mm, scale=0.75]
  \tikzstyle{directed}=[-latex, thick, shorten >=0.5 pt, shorten <=1 pt]
  \tikzstyle{interven}=[directed, densely dotted]
  \node[obs-box] (Y) {$Y$};
  \node[inv-box] (T) [left=0.5cm of Y] {$t$};
  \node[obs-box] (X) [above=0.5cm of T] {$X$};
  \node[lat-exo] (Ux) [left=0.5cm of X] {$U_x$} ;
  \node[lat-exo] (Uy) [above=0.5cm of Y] {$U_y$} ;
  \path  
    (X) edge [directed] (Y)
    (T) edge [directed] (Y)
    (Uy) edge [directed] (Y)
    (Ux) edge [directed] (X)
    ;
\end{tikzpicture}%
      \caption{SCM under $\text{do}(T = t)$}
      \label{fig:scm-treatment-interv-1}
    \end{center}
  \end{subfigure}
  ~
  \begin{subfigure}[b]{\mywidth}
    \begin{center}
      \begin{tikzpicture}[]
  \tikzstyle{lat-exo}=[circle, inner sep=1pt, minimum size = 6.5mm, thick, draw=black!80, node distance = 20mm, scale=0.75]
  \tikzstyle{obs-exo}=[circle, fill=black!20, inner sep=1pt, minimum size = 6.5mm, thick, draw=black!80, node distance = 20mm, scale=0.75]
  \tikzstyle{lat-box}=[rectangle, inner sep=1pt, minimum size = 6mm, draw=black!80, thick, node distance = 20mm, scale=0.75]
  \tikzstyle{obs-box}=[rectangle, fill=black!20, inner sep=1pt, minimum size = 6mm, thick, draw=black!80, node distance = 20mm, scale=0.75]
\tikzstyle{inv-box}=[rectangle, fill=black!20, inner sep=1pt, minimum size = 6mm, thick, draw=black!80,densely dotted, node distance = 20mm, scale=0.75]
  \tikzstyle{directed}=[-latex, thick, shorten >=0.5 pt, shorten <=1 pt]
  \tikzstyle{interven}=[directed, densely dotted]
  \node[obs-box] (Y) {$Y$};
  \node[inv-box] (T) [left=0.5cm of Y] {$T$};
  \node[obs-box] (X) [above=0.5cm of T] {$X$};
  \node[lat-exo] (Ut) [left=0.5cm of T] {$U_t$} ;
  \node[lat-exo] (Ux) [left=0.5cm of X] {$U_x$} ;
  \node[lat-exo] (Uy) [above=0.5cm of Y] {$U_y$} ;
  \path
    (X) edge [directed] (Y)
    (X) edge [interven] (T)
    (T) edge [directed] (Y)
    (Uy) edge [directed] (Y)
    (Ux) edge [directed] (X)
    (Ut) edge [interven] (T)
    ;
\end{tikzpicture}%
      \caption{SCM under $\text{do}(f_T \rightarrow \pi)$}
      \label{fig:scm-treatment-interv-2}
    \end{center}
  \end{subfigure}
  ~
  \caption{
    Treatment model expressed as PGM (\ref{fig:pgm-treatment}), SCM
    (\ref{fig:scm-treatment}),
     and SCM under atomic (\ref{fig:scm-treatment-interv-1}) and policy
     interventions (\ref{fig:scm-treatment-interv-2}).
  }
   \label{fig:treatment}
\end{figure}

There are several ways of encoding causal assumptions in DAG form.
In this paper, 
we focus on 
structural causal models
(SCMs) \citep{pearl2009causal}, which we overview here.\footnote{
Other overviews of various levels of detail can be found elsewhere
\citep{pearl2009causal,madras2019fairness,buesing2019woulda}
.}
SCMs are similar to probabilistic graphical models (PGMs)
\citep{koller2009probabilistic}.
They consist of nodes (random variables representing entities in the world) and
edges (relationships between entities).
However, whereas PGMs only specify a set of conditional independence
relationships, SCMs specify a unique data generating process (analogously, a
\emph{particular} probability factorization, as opposed to the multiple
isomorphic factorizations available in a PGM).

There are two types of nodes in SCMs.
\emph{Endogenous nodes} represent variables of interest within the model, while
\emph{exogenous nodes} are external random variables, representing
the exclusive source of stochasticity induced on the observations (the endogenous nodes).
The edges between nodes are deterministic functions called
\textit{structural equations}.
Hence, a setting of the exogenous nodes corresponds to exactly one setting
of the endogenous nodes.
In Figure \ref{fig:scm-treatment}, the dark squares are endogenous nodes,
representing specific entities such as a credit score, a medical treatment, or a
sensitive attribute.
The light circles are exogenous nodes.
Each endogenous node is the output of a structural equation, e.g. $T =
f_T(X, U_T)$, $Y = f_Y(T, X, U_Y)$.

We can calculate causal quantities under a particular SCM by using the
$do$-operator.
Given the probability distribution implied by the SCM in Figure
\ref{fig:scm-treatment} (call the model $\mathcal{M}$ and the implied joint
distribution $p$), we may wish to ask -- "What would be the expected value of
$Y$ if $T$ were set to 1?"
The corresponding estimand can be denoted $\mathds{E}_{p^{do(T = 1)}}[Y]$.
This differs from the more straightforward conditional probability
$\mathds{E}_{p}[Y | T = 1]$, which describes co-occurences of $Y$ with $T = 1$ in the observed data.
The expression $\mathds{E}_{p^{do(T = 1)}}[Y]$ indicates that expected value of $Y$ is computed under a modified SCM which is specified by $do(T = 1)$; we denote this $\mathcal{M}^{do(T = 1)}$, with the associated probability distribution
$p^{do(T = 1)}$.
$\mathcal{M}^{do(T = 1)}$ is intended to simulate a randomized experiment --- if
the true data-generating process is represented by $\mathcal{M}$, what would
happen to the observed data if we forcibly change the data-generating process,
so that $T = 1$ always?
Graphically, $\mathcal{M}^{do(T = 1)}$ is created by starting with $\mathcal{M}$
(Fig.~\ref{fig:scm-treatment}), removing from the graph all the incoming arrows
to $T$ (in this case, arrows originating from $X$ and $U_{T}$), and setting $T =
1$ (yielding Fig.~\ref{fig:scm-treatment-interv-1}).
This is referred to as an \textit{intervention}.
Under certain conditions \citep{pearl2009causal}, we can identify
$\mathds{E}_{p^{do(T = 1)}}[Y]$ by using observational data generated by $p$ to
simulate sampling from $p^{do(T = 1)}$.
Intervening on the value of $T$ in this way is an \emph{atomic} intervention.

\subsection{Policy Interventions and Off-Policy Evaluation}\label{sec:off-policy-evaluation}
Alternatively, we can intervene directly on the structural equation governing
$T$ (Fig.~\ref{fig:scm-treatment-interv-2}), resulting in model
$\mathcal{M}^{do(f_T \rightarrow \pi)}$ with distribution $p^{do(f_T
\rightarrow \pi)}$.
When an intervention manipulates a structural equation corresponding to a decision maker's policy, we call this a \emph{policy intervention}.
Accordingly, we denote the structural equation under intervention as $\pi$ to emphasize that it represents the decision maker's policy, distinct from the structural $f_T$ present during the previous collection of observational data (which in turn could also be referred to as a policy, say $f_T = \pi_{\text{Hist}}$).

Consider an observational dataset generated by some historical policy $\pi_{\text{Hist}}$.
We may wish to know the expected outcome for some policy $\pi \neq \pi_{\text{Hist}}$, but cannot directly test $\pi$ in the world ourselves.
This \emph{off-policy evaluation} problem
is particularly important in fairness contexts, where running a candidate policy in the world is frequently impossible due to ethical or practical reasons.
In an SCM, off-policy evaluation constitutes estimating expected outcomes under a policy intervention.
In the example from Fig.~\ref{fig:scm-treatment-interv-2}, to estimate the expected value of $Y$ under a new
policy $\pi$, we specify our intervention with $do(f_T \rightarrow \pi)$, and the estimand would be $\mathds{E}_{p^{do(f_T \rightarrow \pi)}}[Y]$.
In general, to denote the expected value of a variable $\mathcal{U}$ under a
target policy $\hat \pi$ which intervenes on a variable $V$, we write
$\mathds{E}_{p^{do(f_V \rightarrow \hat{\pi})}}[\mathcal{U}]$.

\subsection{Benefits of Causal Graphs}
While there are a variety of strategies for modeling in the causal inference literature (the potential outcomes framework of \citet{rubin2005causal} is a popular alternative\footnote{
We note that SWIGs \citep{richardson2013single}, while not the focus of this work, provide a graphical method to express potential outcomes that could also used to study long-term fairness.
}), we believe that causal \emph{graphs} as pioneered by \cite{pearl2009causal} convey several benefits of particular interest in applications with fairness concerns.
We outline these benefits below.
\paragraph{Visualization} Many problems in long-term fairness have a large number of variables, and require collaboration across disciplines and with policy makers or regulators.
Graphical structure allows for mathematical manipulation of many variables, and can convey basic assumptions to non-technical stakeholders.

\paragraph{Introspection}
Using causal language to be explicit about assumptions is useful for learning better policies (we discuss one such example in Sec.~\ref{sec:off-policy-eval}).
Using a graph to convey the causal assumptions is a stylistic choice, but it facilitates the interrogation of complex assumptions (with many variables).
Since the usefulness of causal inferences often rests assumptions that cannot be readily tested, it is especially important to hold these assumptions to a high degree of scrutiny; the use of graphs to convey causal assumptions could empower non-technical stakeholders to participate in this process.

\paragraph{Evaluation} Causal graphs convey a number of methodological benefits, especially in improving off-policy evaluation (Sec.~\ref{sec:off-policy-eval}), enabling expressive simulation, and suggesting relevant sensitivity analyses (Sec.~\ref{sec:individual-dynamics}).
Furthermore, encoding causal assumptions using graphical language exposes an underlying \emph{computation graph}.
Under mild assumptions, the topology of a computation graph can be used to programatically derive a large family of estimators for use in off-policy evaluation and gradient-based policy learning \citep{schulman2015gradient,weber2019credit}.
In the context of causal inference, graph topology can assist in determining the identifiability of policy interventions from observational data; see discussion of ``dynamic treatment regimes'' by \citet{hernan2020causal} for further detail.


\section{Causal Interpretations of Dynamic Fairness Models} \label{sec:motivating-dags}
In this section, we demonstrate how SCMs present a unifying framework for the literature on fairness in dynamical systems.
We focus on how causal mechanisms enable easier explication of underlying modeling assumptions, yielding insight into the component parts of the model, types of bias which could arise, and the effects of hypothetical interventions.
Our aim is not to promote a particular dynamical model or fairness objective/constraint, either in general or for specific problem domains; rather we aim to provide a tool with which policymakers and practitioners alike can analyze a long-term unfairness problem.
We discuss SCM formulations of three models of fairness in dynamical systems (see Appendix~\ref{sec:other-scms} for several
more examples):
\begin{enumerate}
    \item Fair-MDP: a motivating example showing how bias can arise in a generic sequential decision process.
    \item Lending: \citet{liu2018delayed}'s single-step model of a loan application.
    \item Repeated classification: \citet{hashimoto2018fairness}'s model of the dynamics of a changing population's preferences with unobserved sensitive attributes.
\end{enumerate}{}

\begin{figure}[t!]
\centering
\noindent\resizebox{.3\textwidth}{!}{
  \newcommand{\ssi}{2}  %
\newcommand{\exu}{.3cm}  %
\newcommand{\nodescale}{.65}  %
\begin{tikzpicture}[scale=0.7, >=stealth]
  \tikzstyle{lat-exo}=[circle, inner sep=1pt, minimum size = 6.5mm, thick, draw=black!80, node distance = 20mm, scale=\nodescale, font=\large]
  \tikzstyle{obs-exo}=[circle, fill=black!20, inner sep=1pt, minimum size = 6.5mm, thick, draw=black!80, node distance = 20mm, scale=\nodescale]
  \tikzstyle{lat-box}=[rectangle, inner sep=1pt, minimum size = 6mm, draw=black!80, thick, node distance = 20mm, scale=\nodescale, font=\large]
  \tikzstyle{obs-box}=[rectangle, fill=black!20, inner sep=1pt, minimum size = 6mm, thick, draw=black!80, node distance = 20mm, scale=\nodescale, font=\large]
\tikzstyle{inv-box}=[rectangle, fill=black!20, inner sep=1pt, minimum size = 6mm, thick, draw=black!80,densely dotted, node distance = 20mm, scale=\nodescale]
  \tikzstyle{directed}=[-latex, thick, shorten >=0.5 pt, shorten <=1 pt]
  \tikzstyle{interven}=[directed, densely dotted]
  \tikzstyle{empty}=[]

\node[obs-box] (a)                              at (-1,4)         {$A$};
\node[lat-exo]  (ua)              [above left=\exu of a]     {$U_{A}$};
\node[obs-box]  (x0)                          at (1,2.5)       {$X^0$};
\node[lat-exo] (ux0)                  [above=\exu of x0]   {$U_{X^0}$};
\node[obs-box]  (t0)                            at (2,1)       {$T^0$};
\node[lat-exo] (ut0)                   [left=\exu of t0]   {$U_{T^0}$};
\node[obs-box]  (x1)                     at (1+\ssi,2.5)       {$X^1$};
\node[lat-exo] (ux1)                  [above=\exu of x1]   {$U_{X^1}$};
\node[obs-box]  (t1)                       at (2+\ssi,1)       {$T^1$};
\node[lat-exo] (ut1)                   [left=\exu of t1]   {$U_{T^1}$};
\node[obs-box]  (x2)                   at (1+2*\ssi,2.5)       {$X^2$};
\node[lat-exo] (ux2)                  [above=\exu of x2]   {$U_{X^2}$};
\path
         (ua) edge [directed]  (a)
        (ux0) edge [directed] (x0)
         (x0) edge [directed] (t0)
          (a) edge [directed              ] (x0)
          (a) edge [directed, bend left=40] (x1)  
          (a) edge [directed, bend left=50] (x2)  
          (a) edge [directed, bend right=10] (t0)  
          (a) edge [directed, bend right=100] (t1)  
        (ut0) edge [directed] (t0)
        (x0)  edge [directed] (x1)
        (t0)  edge [directed] (x1)
        (ux1) edge [directed] (x1)
         (x1) edge [directed] (t1)
        (ut1) edge [directed] (t1)
        (x1)  edge [directed] (x2)
        (t1)  edge [directed] (x2)
        (ux2) edge [directed] (x2)
        ;
\node[empty]   (ea)              [below=\exu of a] {};
\end{tikzpicture}
}
\caption{
  The Fair-MDP: a motivating example model for estimating questions of fairness in sequential decision-making problems.
  }\label{fig:motivating-example}
\end{figure}
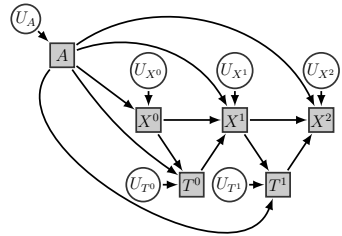

\subsection{Fair-MDP: A Motivating Example}
We begin by suggesting a minimal characterization of a sequential decision process in the fairness setting.
Consider the following SCM (see Fig. \ref{fig:motivating-example}), with the factorization:
\begin{equation}
    \begin{aligned}
        A &= f_A(U_A)\\
        X^0 &= f_{X^0}(A, U_{X^0})\\
        T^{k} &= f_T(X^k, A, U_{T^k}), k = 0 \dots K\\
        X^{k + 1} &= f_X(X^k, T^{k}, A, U_{X^{k + 1}}), k = 0 \dots K\\
    \end{aligned}{}
\end{equation}{}
This is similar to a Markov Decision Process (MDP), in that the key elements are states ($X$), actions ($T$), a policy ($f_T$) and a transition function ($f_X$).
However, we note that it is not fully Markovian --- the sensitive attribute persists across states, affecting all aspects of the problem.
This aligns with standard fairness intuitions, since the sensitive attribute is generally considered to be somewhat holistic and immutable by $T$.
We denote this model the \textit{Fair-MDP}, since it becomes an MDP when we condition on the sensitive attribute, and the inclusion of this attribute permits fairness considerations.\footnote{
In general, MDPs are typically used to define all relevant variables as part of a “state”, and methods from this literature can be applied to long term fairness problems. 
Causal graphs enable fine-grained modeling of the dynamics with a state (see Fig.~\ref{fig:scm-hu} in Appendix~\ref{sec:other-scms} for one such example), which can be practically useful in long-term fairness problems, e.g. in improving off-policy evaluation (Sec.~\ref{sec:off-policy-eval}).
}
We can think of $X$ as some feature of an individual, which our policy is aiming to maximize, and consider the final $X^k$ in the sequence as the reward.

We can use this model to examine different fairness issues in the sequential setting.
For instance, consider the issue of feeback loops \citep{ensign2018runaway,lum2016predict}.
Suppose that the initial feature distribution $P(X^0 | A)$ is uneven: $\mathds{E}[X^0 | A = 1] > \mathds{E}[X^0 | A = 0]$.
Additionally, suppose a threshold policy is applied, with 
${T^k = f_T(X^k, \cdot, \cdot) = \ind{X^k > \tau}}$
and that the application of the treatment causes $X$ to increase: ${\mathds{E}(X^{k + 1} | X^k, T^k = 1) > X^k}$ (and $T^k = 0$ causes the opposite effect).
Then, we might expect to see a \textit{feedback loop}, as observed in \citet{ensign2018runaway}, where one group's average reward increases continuously over time, and the other group's decreases.

Off-policy estimation for a policy $\pi$ in this model amounts to estimating $\mathds{E}_{p^{do(f_T \rightarrow \pi)}}[X^K]$.
We note that this is a non-trivial problem --- if we only observe data generated by some historical policy $\pi_H \neq \pi$, then the values of $X^K$ under the actions that $\pi$ would have taken may not be available in our data.
In this case, the naive estimator $\mathds{E}_{p}[X^K]$ will be biased.
We return to the off-policy estimation question in Section \ref{sec:off-policy-eval}, with a causal approach.

\subsection{Lending}\label{sec:lending-overview}
We turn to the model from \citet{liu2018delayed}, which examines
threshold-based classification in general, but with specific focus on the
lending setting.
Our SCM formulation of this model can be seen in Figure~\ref{fig:scm-liu}.
In this model, a person with group membership (a.k.a. sensitive attribute) $A$
receives a credit score $X$, and applies to a bank for a loan.
The bank makes a binary decision $T$ about whether to award the loan using the policy $f_T$.
The binary potential outcome $Y$ is realized, which is converted to
institutional profit or loss only if $T=1$.\footnote{
  Therefore this model does not capture a notion of opportunity loss for
  \emph{not} extending a loan to applicants who \emph{are} qualified.
}
Finally, the applicant's credit score is modified to $\tilde X$ (increased on repayment, decreased on default, static if $T=0$).\footnote{
  Likewise, the applicant's score does not change in the absence of a loan; this
  assumption may be inaccurate, since \textit{not} receiving a loan could create
  additional financial issues for the applicant.
}
The bank's utility is measured through their profit $\mathcal{U}$ (a sum over the
individual profits $u$) as well as the expected score change $\Delta_j$,
representing the average change in credit score after one time-step among
members of group $A = j$.
Varying the loan policy can achieve different values of $\mathcal{U}, \Delta_j$, resulting in outcomes with different fairness properties.

\citet{liu2018delayed} consider the effect of various 
threshold policies
for loan assignment under this model, namely the expected values of
$\mathcal{U}$ and $\Delta$ for some policies with group-specific thresholds $\tau \triangleq
(\tau_0, \tau_1)$ that offer loans to applicants of group $j$ with score $X$ if
and only if their credit score $X > \tau_j$.
They show that different thresholds satisfy different criteria: maximum profit
(\MaxProf), demographic parity (\DP), and equal opportunity (\EqOpp).
In the language of our paper, comparing threshold policies is done through policy evaluation and intervention.
Denoting by $\pi_{\tau}$ a threshold 
policy per group
$\tau$,
these results can be phrased with the tool of policy
intervention: we evaluate the policy $\pi_{\tau}$ by estimating
the quantities $\mathds{E}_{p^{do(f_T \rightarrow \pi_{\tau})}} [\mathcal{U}]$
and $\mathds{E}_{p^{do(f_T \rightarrow \pi_{\tau})}} [\Delta_j] \ \forall j$,
for various $\tau$ computed under different fairness criteria.
We discuss off-policy evaluation in this model in Section \ref{sec:off-policy-eval}.

This SCM interpretation suggests several
potential extensions, such as 
evaluating outcomes over
multiple steps or adding extra actors to the model.
We discuss these in detail in Section \ref{sec:individual-dynamics}, where we
provide a case study of this SCM.

\subsection{Repeated Classification}

\begin{figure}[b!]
\noindent\resizebox{.5\textwidth}{!}{
  \newcommand{\stepsize}{6}
\begin{tikzpicture}[scale=1., >=stealth]
\tikzstyle{empty}=[]
\tikzstyle{lat-exo}=[circle, inner sep=1pt, minimum size = 6.5mm, thick, draw=black!80, node distance = 20mm, scale=0.75, font=\huge]
\tikzstyle{obs-exo}=[circle, fill=black!20, inner sep=1pt, minimum size = 6.5mm, thick, draw=black!80, node distance = 20mm, scale=0.75, font=\huge]
\tikzstyle{lat-box}=[rectangle, inner sep=1pt, minimum size = 6mm, draw=black!80, thick, node distance = 20mm, scale=0.75, font=\huge]
\tikzstyle{obs-box}=[rectangle, fill=black!20, inner sep=1pt, minimum size = 6mm, thick, draw=black!80, node distance = 20mm, scale=0.75, font=\huge]
\tikzstyle{inv-box}=[rectangle, fill=black!20, inner sep=1pt, minimum size = 6mm, thick, draw=black!80,densely dotted, node distance = 20mm, scale=0.75, font=\huge]
\tikzstyle{connect}=[-latex, thick]
\tikzstyle{undir}=[thick]
\tikzstyle{directed}=[->, thick, shorten >=0.5 pt, shorten <=1 pt]
\tikzstyle{semi-box}=[rectangle with diagonal fill, diagonal top color=black!20, diagonal bottom color=white, diagonal from left to right, draw]
\tikzstyle{interven}=[directed, densely dotted]

\node[obs-box] at (4,5)      (n0) {$N^0$};
\node[lat-exo] at (2,5)     (bk0) {$b_k$};
\node[lat-box] at (2,4)     (lk0) {$\lambda_k^0$};
\node[lat-box] at (2,3)     (ak0) {$\alpha_k^0$};
\node[lat-exo] at (1,3)     (pk0) {$P_k$};
\node[lat-box] at (3,3)     (rk0) {$R_k^0$};
\node[lat-box] at (2,2)      (z0) {$Z^0$};
\node[obs-box] at (1,0.5)    (x0) {$X^0$};
\node[obs-box] at (2,0.5)    (y0) {$Y^0$};
\node[obs-exo] at (1,-1)   (u0) {$U_{\theta}^0$};
\node[inv-box] at (2,-1)   (t0) {$\theta^0$};
\node[obs-box] at (3,0.5)   (yh0) {$\hat Y^0$};
\node[obs-box] at (4+\stepsize,5)      (n1) {$N^1$};
\node[lat-exo] at (2+\stepsize,5)     (bk1) {$b_k$};
\node[lat-box] at (2+\stepsize,4)     (lk1) {$\lambda_k^1$};
\node[lat-box] at (2+\stepsize,3)     (ak1) {$\alpha_k^1$};
\node[lat-exo] at (1+\stepsize,3)     (pk1) {$P_k$};
\node[lat-box] at (3+\stepsize,3)     (rk1) {$R_k^1$};
\node[lat-box] at (2+\stepsize,2)      (z1) {$Z^1$};
\node[obs-box] at (1+\stepsize,0.5)    (x1) {$X^1$};
\node[obs-box] at (2+\stepsize,0.5)    (y1) {$Y^1$};
\node[obs-exo] at (1+\stepsize,-1)   (u1) {$U_{\theta}^1$};
\node[inv-box] at (2+\stepsize,-1)   (t1) {$\theta^1$};
\node[obs-box] at (3+\stepsize,0.5)   (yh1) {$\hat Y^1$};
\node[empty] at (1+1.75*\stepsize,4) (lk2a) {};
\node[empty] at (1+1.75*\stepsize,3.5) (lk2b) {};
  \path   
        (bk0) edge [directed]                (lk0)
        (lk0) edge [directed, bend left=25]   (n0)
        (lk0) edge [directed]                (ak0)
        (ak0) edge [directed]                 (z0)
         (z0) edge [directed]                 (x0)
         (z0) edge [directed]                 (y0)
        (pk0) edge [directed]                 (x0)
        (pk0) edge [directed]                 (y0)
         (x0) edge [interven]                 (t0)
         (y0) edge [interven]                 (t0)
         (x0) edge [directed, bend right=45] (yh0)
         (t0) edge [directed]                (yh0)
         (u0) edge [directed]                 (t0)
         (y0) edge [directed]                (rk0)
         (yh0) edge [directed]               (rk0)
         (z0) edge [directed]                (rk0)         
        (lk0) edge [directed]  (lk1)
        (rk0) edge [directed]  (lk1)
        (bk1) edge [directed]                (lk1)
        (lk1) edge [directed, bend left=25]   (n1)
        (lk1) edge [directed]                (ak1)
        (ak1) edge [directed]                 (z1)
         (z1) edge [directed]                 (x1)
         (z1) edge [directed]                 (y1)
        (pk1) edge [directed]                 (x1)
        (pk1) edge [directed]                 (y1)
         (x1) edge [interven]                 (t1)
         (y1) edge [interven]                 (t1)
         (x1) edge [directed, bend right=45] (yh1)
         (t1) edge [directed]                (yh1)
         (u1) edge [directed]                 (t1)
         (y1) edge [directed]                (rk1)
        (yh1) edge [directed]                (rk1)
        (z1) edge [directed]                 (rk1)         
        (lk1) edge [-, thick] (lk2a)
        (rk1) edge [-, thick] (lk2b)
        ;
\plate {platek0} {(lk0)(ak0)(bk0)(rk0)(pk0)(z0)} {$K$} ;
\plate {platen0} {(z0)(x0.north)(y0)(yh0)(platek0.south west)(platek0.south east)} {$N^0$} ;
\plate {platek1} {(lk1)(ak1)(bk1)(rk1)(pk1)(z1)} {$K$} ;
\plate {platen1} {(z1)(x1.north)(y1)(yh1)(platek1.south west)(platek1.south east)} {$N^1$} ;
\end{tikzpicture}
}
\caption{
    SCM for the repeated loss minimization model discussed by
      \citet{hashimoto2018fairness}.
    The dotted arrows highlight the policy as the learning algorithm that produces parameters $\theta^t$, which in turn affect the predictions $\hat Y_i^t$.
    See Table \ref{tab:scm-hashimoto} for explanation of all symbols and text for description.
  }\label{fig:scm-hashimoto}
\end{figure}

Finally, we examine the repeated classification setting discussed by
\citet{hashimoto2018fairness}, presented in SCM form in
Figure~\ref{fig:scm-hashimoto}.
The model is fairly general, and the authors discuss several domains where it
could apply (e.g. speech recognition, text auto-completion).
A binary classifier with parameters $\theta$
is repeatedly trained on a population of individuals with features $X$ and labels $Y$.
The population distribution is a mixture of components $P=\sum_k \alpha_k P_k$,
where each of the $k$ demographic groups has proportion $\alpha_k$ 
(with ${\sum_k \alpha_k = 1}$)
and a unique distribution over the input-output pairs $P_k(X, Y)$.
Group memberships (i.e. cluster assignments) $Z \in [1\ldots k]$ are not
observed.

The key idea is that the group distributions $P_k$ remain static over time, but
their relative proportions $\alpha_k$ change dynamically in response to the
classifier performance on the $k$-th group.
At the $t$-th step, the classifier is trained on the overall population
$\{(X^t_i,Y^t_i)\}$, yielding classifier parameter $\theta^t$ and
predictions\footnote{
  Using held-out data for the predictions is expressible via a small change to
  the SCM.
} $\hat{Y}^t$.
At each step, some subjects choose to stay in the population, some choose to
leave, and some new subjects are added to the pool.
In particular, the Poisson parameter $\lambda_k$ (proportional to mixing
coefficient $\alpha_k$) is computed as a function of the per-group risk $R_k$.
Misclassified subjects are more likely to leave, so under-served groups shrink
over time.
The authors coin this phenomenon as \textit{disparity amplification}.
Interestingly, disparity amplification can \emph{improve} the overall
loss/accuracy since the shrinking minority group (whose accuracy may be decreasing) contributes less to these
global metrics as time proceeds.
To mitigate disparity amplification, \citet{hashimoto2018fairness} propose a
robust optimization technique that seeks low loss for worst-case group
assignments $Z$ (assuming a minimum group size).

This SCM suggests several interesting interventions:
\begin{enumerate}
    \item Intervention on \emph{latent dynamics}: $\text{do}(f_\lambda
      \rightarrow \hat f_\lambda)$ represents an intervention on population dynamics,
      which we could use to test how policies affect the entry and exit of various groups from the environment over time.
      $\text{do}(b_k=\hat b_k)$ is a simple atomic intervention of a similar
      flavor, which changes the expected number of individuals entering each
      group at a given time step.
    \item Intervention on \emph{group distributions}: $\text{do}(P_k \rightarrow
      \hat P_k)$ 
      shifts the distribution over input-output pairs for group $k$,
      which could be carried out at one or every time step.
\end{enumerate}

We do not present experiments on this model,
but include it to suggest the types of analyses and extensions possible for SCMs with increased complexity.
See Appendix \ref{sec:other-scms} for more sophisticated models from the fairness in dynamical systems literature represented as SCMs.


\section{Related Work}\label{sec:related-work}
\paragraph{Dynamical Fairness}
There has been work on modeling the long-term dynamics of fairness in a range
of potential domains.
Recently, the first paper to bring these issues to light was
\citet{lum2016predict}, discussing the bias feedback loops which could arise
in predictive policing systems, with follow-up work by
\citet{ensign2018runaway}.
Domains such as hiring \citep{hu2018short}, loans \citep{mouzannar2019fair}, and
recommender systems \citep{hashimoto2018fairness,bountouridis2019siren} have
also been explored in this way.
Other related explorations have dealt with short-term dynamics
\citep{liu2018delayed} and strategic actions
\citep{hu2019disparate,milli2018social}.
There is also a line of work studying the long-term effects of affirmative
action, with some classic works from the economics literature
\citep{coate1993will,foster1992economic}, and more recent computer science
focused work \citep{kannan2019downstream}.
On the theoretical side, several general algorithms for improved fairness in
sequential decision-making have been characterized, with work discussing bandits
\citep{joseph2016fairness}, reinforcement learning \citep{jabbari2017fairness},
and importance sampling estimators \citep{doroudi2017importance}.
The work of \citet{d2020fairness}---which most closely relates to ours---studies
long-term outcomes for existing fair ML methods, emphasizing \emph{agents}
and \emph{environments} as modeling primitives.
Our contributions can be seen as complementary, emphasizing the role of \emph{causal}
modeling primitives within a dynamical system, both in terms of estimation from observational
trajectories, and building expressive simulators for evaluating agents and environments.

\paragraph{Causality}
Causal modeling has been used in a variety of non-dynamic fair machine learning
approaches.
Work on counterfactual fairness \citep{kusner2017counterfactual} has considered
fairness definitions which encourage models to treat examples similarly to
hypothetical situations where they were from the other group.
Some other works focus on learning fair policies from biased observational data
\citep{madras2019fairness,kusner2019making} or on learning decision
rules which follow only causal paths deemed to be non-discriminatory
\citep{kilbertus2017avoiding,nabi2018fair,nabi2019learning}.
Another line of work interprets previously proposed fairness criteria from a causal perspective \cite{zhang2018equality,zhang2018fairness}.

Outside of fairness, \citet{everitt2019understanding} propose using influence
diagrams as a framework for understanding safety in AI systems.

\begin{figure}[hb!]
  \centering
  \begin{subfigure}[t]{0.23\textwidth}
    \noindent\resizebox{\textwidth}{!}{
      \begin{tikzpicture}[scale=0.85, >=stealth]
  \tikzstyle{lat-exo}=[circle, inner sep=1pt, minimum size = 6.5mm, thick, draw=black!80, node distance = 20mm, scale=0.75]
  \tikzstyle{obs-exo}=[circle, fill=black!20, inner sep=1pt, minimum size = 6.5mm, thick, draw=black!80, node distance = 20mm, scale=0.75]
  \tikzstyle{lat-box}=[rectangle, inner sep=1pt, minimum size = 6mm, draw=black!80, thick, node distance = 20mm, scale=0.75]
  \tikzstyle{obs-box}=[rectangle, fill=black!20, inner sep=1pt, minimum size = 6mm, thick, draw=black!80, node distance = 20mm, scale=0.75, font=\huge]
\tikzstyle{inv-box}=[rectangle, fill=black!20, inner sep=1pt, minimum size = 6mm, thick, draw=black!80,densely dotted, node distance = 20mm, scale=0.75, font=\huge]
  \tikzstyle{directed}=[-latex, thick, shorten >=0.5 pt, shorten <=1 pt]
  \tikzstyle{interven}=[directed, densely dotted]
  \tikzstyle{empty}=[]

\newcommand{\exounit}{.3cm}
\node[obs-box] at (1,3)  (a) {$A$};
\node[lat-exo] (ua) [above left=\exounit of a] {$U_{A}$};
\node[obs-box] at (2,4)  (x) {$X$};
\node[lat-exo] (ux) [above right=\exounit of x] {$U_{X}$};
\node[obs-box] at (2,2)  (t) {$T$};
\node[lat-exo] (ut) [below left=\exounit of t] {$U_{T}$};
\node[obs-box] at (3,3)  (y) {$Y$};
\node[lat-exo] (uy) [above right=\exounit of y] {$U_{Y}$};
\node[obs-box] at (6,2)  (U) {$\mathcal{U}$};
\node[obs-box] at (4.5,2)  (u) {$u$};
\node[obs-box] at (4.5,3)  (xt) {$\tilde X$};
\node[obs-box] at (5.5,7)  (Dj) {$\Delta$};
\node[empty] (e) [below right=\exounit of Dj] {};
\path   (ua) edge [directed] (a)
        (a)  edge [directed] (x)
        (ux) edge [directed] (x)
        (x)  edge [directed] (t)
        (a)  edge [directed] (t)
        (a)  edge [directed] (y)
        (ut) edge [directed] (t)
        (x)  edge [directed] (y)
        (uy) edge [directed] (y)
        (y)  edge [directed] (u)
        (t)  edge [directed] (u)
        (y)  edge [directed] (xt)
        (t)  edge [directed] (xt)
        (x)  edge [directed, bend left=55] (xt)
        (u)  edge [directed] (U)
        (xt) edge [directed, bend right=10] (Dj)
        (x)  edge [directed, bend left=40] (Dj)
        (a)  edge [directed, bend left=45] (Dj)
        ;
\plate {} {(ua)(a)(x)(ux)(t)(ut.north west)(y)(uy)(u)(xt)} {$N$} ;
\plate {} {(Dj)(e)} {$|\mathcal{A}|$} ;
\end{tikzpicture}
    }
  \caption{.
    Our SCM formulation of the one-step dynamics.
  }\label{fig:scm-liu}
  \end{subfigure}%
  \hfill
  \begin{subfigure}[t]{0.23\textwidth}
    \noindent\resizebox{\textwidth}{!}{
      \begin{tikzpicture}[scale=0.85, >=stealth]
  \tikzstyle{lat-exo}=[circle, inner sep=1pt, minimum size = 6.5mm, thick, draw=black!80, node distance = 20mm, scale=0.75]
  \tikzstyle{obs-exo}=[circle, fill=black!20, inner sep=1pt, minimum size = 6.5mm, thick, draw=black!80, node distance = 20mm, scale=0.75]
  \tikzstyle{lat-box}=[rectangle, inner sep=1pt, minimum size = 6mm, draw=black!80, thick, node distance = 20mm, scale=0.75]
  \tikzstyle{obs-box}=[rectangle, fill=black!20, inner sep=1pt, minimum size = 6mm, thick, draw=black!80, node distance = 20mm, scale=0.75, font=\huge]
\tikzstyle{inv-box}=[rectangle, fill=black!20, inner sep=1pt, minimum size = 6mm, thick, draw=black!80,densely dotted, node distance = 20mm, scale=0.75, font=\huge]
  \tikzstyle{directed}=[-latex, thick, shorten >=0.5 pt, shorten <=1 pt]
  \tikzstyle{interven}=[directed, densely dotted]
  \tikzstyle{empty}=[]

\newcommand{\exounit}{.3cm}
\node[obs-box] at (1,3.5)  (a) {$A$};
\node[lat-exo] (ua) [above left=\exounit of a] {$U_{A}$};
\node[obs-box] at (2,4.5)  (x) {$X$};
\node[lat-exo] (ux) [above right=\exounit of x] {$U_{X}$};
\node[inv-box] at (2,3.4)  (hx) {$\hat X$};
\node[obs-box] at (2,2)  (t) {$T$};
\node[lat-exo] (ut) [below left=\exounit of t] {$U_{T}$};
\node[obs-box] at (3,3.5)  (y) {$Y$};
\node[lat-exo] (uy) [above right=\exounit of y] {$U_{Y}$};
\node[obs-box] at (6,2)  (U) {$\mathcal{U}$};
\node[obs-box] at (4.5,2)  (u) {$u$};
\node[obs-box] at (4.5,3.5)  (xt) {$\tilde X$};
\node[obs-box] at (5.5,7)  (Dj) {$\Delta$};
\node[empty] (e) [below right=\exounit of Dj] {};
\path   (ua) edge [directed] (a)
        (a)  edge [directed] (x)
        (ux) edge [directed] (x)
        (x)  edge [interven] (hx)
        (hx) edge [directed] (t)
        (a)  edge [directed] (t)
        (a)  edge [directed, bend right=45] (y)
        (ut) edge [directed] (t)
        (x)  edge [directed] (y)
        (uy) edge [directed] (y)
        (y)  edge [directed] (u)
        (t)  edge [directed] (u)
        (y)  edge [directed] (xt)
        (t)  edge [directed] (xt)
        (x)  edge [directed, bend left=55] (xt)
        (u)  edge [directed] (U)
        (xt) edge [directed, bend right=10] (Dj)
        (x)  edge [directed, bend left=40] (Dj)
        (a)  edge [directed, bend left=45] (Dj)
        ;
\plate {} {(ua)(a)(x)(ux)(t)(ut.north west)(y)(uy)(u)(xt)} {$N$} ;
\plate {} {(Dj)(e)} {$|\mathcal{A}|$} ;
\end{tikzpicture}
    }
  \caption{
  An extension emphasizing
    the role of the credit bureau. 
  }\label{fig:scm-liu-extras}
  \end{subfigure}%
  \caption{
    Causal interpretation of lending scenario from 
    \citet{liu2018delayed}.
    See Sections \ref{sec:lending-overview}, \ref{sec:off-policy-eval} and \ref{sec:individual-dynamics}
    for discussion, and Table~\ref{tab:scm-liu} in
    Appendix~\ref{sec:symbol-legends} for symbol legend.
  }\label{fig:scm-liu-both}
\end{figure}
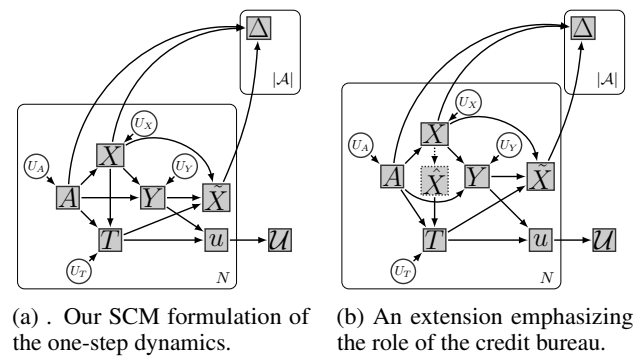
\section{Off-Policy Evaluation and Selection}\label{sec:off-policy-eval}
Given historical observations, how can we estimate the real-world impact of deploying a \emph{new} policy (e.g. one that incorporates fairness constraints)?
This question motivates one of the key tasks required for improving long-term ML
fairness: \emph{off-policy evaluation}.
As noted in Sec. \ref{sec:background}, here we must rely on \emph{observational data}
since it is often unethical or unsafe to test candidate policies in the world (e.g. an A/B test).
In this section, we demonstrate empirically that causal reasoning improves
off-policy evaluation from observational data.

In this experiment, we consider a scenario where the bank has historical data from a profit-maximizing policy (\MaxProf)  and wishes to learn and estimate the quality of an equal opportunity policy (\EqOpp) before deploying it (the off-policy estimation/learning problem).
We use the lending setting of \citet{liu2018delayed} under our SCM interpretation\footnote{
Our aim in this Section is to demonstrate the additional capabilities of causal modeling in this context, and not to adjudicate on whether this choice of dynamics model is correct or appropriate to the lending setting.
In decision making problems with material consequences for individuals and groups (such as lending), care should be taken when incorporating sensitive group information into a causal model.
For example, the generative process studied here includes a causal link from sensitive attribute (which represents race) to credit score.
Although this is consistent with the original paper and captures the statistics of the dataset, it could also be subject to criticism around modeling race as a cause of social position rather than being socially ascribed; see \citet{benthall2019racial} and \citet{hanna2020towards} for further discussion.
} (see Figure \ref{fig:scm-liu} for depiction and Appendix \ref{app:scm-liu} for full specification).
The key (non-trivial) structural equations of the SCM are:
\begin{equation}
    \begin{aligned}
        X &= f_X(A, U_X)\\
        T &= f_T(X, A, U_T)\\
        Y &= f_Y(X, A, U_Y)\\
    \end{aligned}
\end{equation}
which are the feature distribution, the historical treatment policy, and the outcome distribution, respectively.
The change in individual score $c$, the bank's utility $u$, and the next-step score $\tilde{X}$, are simple functions 
of the other variables: $(c, u) = (c_+, u_+)$ if $Y = 1$ or $(c_-, u_-)$ if $Y = 0$, 
and $\tilde{X} = X + c$ (for constants $c_+, u_+ > 0; c_-, u_- < 0$).
As in \citet{liu2018delayed}, we focus on threshold policies, which are defined by group-specific thresholds $\tau \triangleq (\tau_0, \tau_1)$ that offer loans to applicants of group $j$ with score $X$ if
and only if their credit score $X > \tau_j$.

\subsection{Procedure}
In order to compute good thresholds $\tau_j$ for various lending policies (maximum profit, equal opportunity, etc.), 
\citet{liu2018delayed} make a very strong assumption in their method: that these underlying dynamics parameters ($f_X, f_T, f_Y, c_+, c_-, u_+, u_-$) of the system are known.
This is stronger than just assuming the causal structure, as we do in Fig. \ref{fig:scm-liu}.
The causal structure implies the general functional form for the data generating process.
However, \citet{liu2018delayed} assume not just the form but that the function parameter values are known.
In practice, these functions will rarely be known, and must be estimated from observational data.
Therefore any off-policy selection or learning hinges on the quality of these estimates.

Some of these unknown parameters (e.g. $u_+, u_-, f_T$) are easy to estimate from data.
However, one in particular is difficult: the outcome function $Y = f_Y(X, A, U_Y)$.
To understand why estimating $f_Y$ from data is difficult, we must note that $Y$ is a \textit{causal} quantity.
Specifically, $Y$ is a \emph{potential outcome} \citep{rubin2005causal}: it is the probability of a person repaying a loan \textit{were they to receive one}.\footnote{
Using the notation of \citet{rubin2005causal}, we could denote it as $Y_1$.
}
Estimating $Y$ is difficult because it is often missing: we only observe $Y$ when a loan was given in the observational data.
Therefore, straightforward estimates may be biased or high variance.

This difficulty of estimating $Y$ propagates into the rest of the problem;
$u$ and $\Delta$ have the same issues: they are potential outcomes, only observed when the treatment is given ($T = 1$).
Therefore, choosing the policy thresholds---which involves estimating $(u, \Delta)$---is inherently a causal problem.

Given a policy $\pi$, we focus on computing an off-policy estimator ${\mathcal{E}(\pi) \approx \mathds{E}_{p^{do(f_T \rightarrow \pi)}}[u]}$.
A simple estimator can be derived via regression: first learn a function to approximate ${f_\text{Reg}(X, A) \approx {\mathds{E}_{p^\text{obs}}[u | X, A]}}$ in the observational data; then apply this regression for every individual where $\pi$ suggests giving the treatment: ${\mathcal{E}_{Reg}(\pi) = \E_{p^\text{obs}(X,A)}[f_\text{Reg}(X,A)|\pi(X,A)=1]}$.
This is a natural baseline in the absence of causal reasoning.

However, we can further improve this estimator.
As noted previously, $u$ is missing from the observational data in a biased way.
Therefore, we can approach the off-policy estimation problem as a missing data problem --- an area for which causal inference has developed a number of tools.
Crucially, the set $\{X, A\}$ satisfies the \emph{backdoor criterion} from $T$ to $u$ in the SCM (see Fig. \ref{fig:scm-liu}).
This justifies\footnote{
We also rely on the assumptions of \emph{overlap} (in this case requiring a mildly stochastic historical policy) and \emph{consistency}.
While these are typical in the causal inference literature, they may be difficult to verify from observational data in some settings.
See Appendices \ref{app:ope-details} and \ref{app:scm-liu} for further discussion.
} the use of a doubly robust estimator as presented by \citet{zhang2012robust}, an estimator that 
combines a regression-based and an inverse-propensity estimator \citep{bang2005doubly} to reduce bias and variance.\footnote{
The doubly robust estimator can also be interpreted as applying the regression estimator as a control variate to the importance sampling estimator; see \citet{dudik2011doubly} for discussion.
}
With $C_i = \mathbb{1}[\pi(X_i, A_i) = T]$, the estimator is
\begin{equation*}
    \begin{aligned}
        \mathcal{E}_{DR} &= \frac{1}{n} \sum_{i=1}^n \Big[ \frac{C_i(\pi) u_i}{P(C_i(\pi) = 1 | X_i, A_i)} \\
        &- \frac{C_i(\pi) - P(C_i(\pi) = 1 | X_i, A_i)}{P(C_i(\pi) = 1 | X_i, A_i)}
        f_{Reg}(X_i, A_i) \Big].
    \end{aligned}
\end{equation*}{}
We can use an analogous estimator for $\Delta$, where the same backdoor criterion holds.
To summarize, we took the following steps to derive the causal estimator:
\begin{enumerate}
    \item We first recognized that $u$ was implicitly a causal quantity (a potential outcome).
    \item We next examined the SCM to identify a set of variables which satisfied the backdoor criterion between $T$ and $u$.
    \item We finally applied a specialized estimator from the causal inference literature to improve performance.
\end{enumerate}{}

\subsection{Results}
\begin{figure}[b!]
  \centering
  \begin{subfigure}[t]{0.23\textwidth}
    \noindent\resizebox{\textwidth}{!}{
      \includegraphics[]{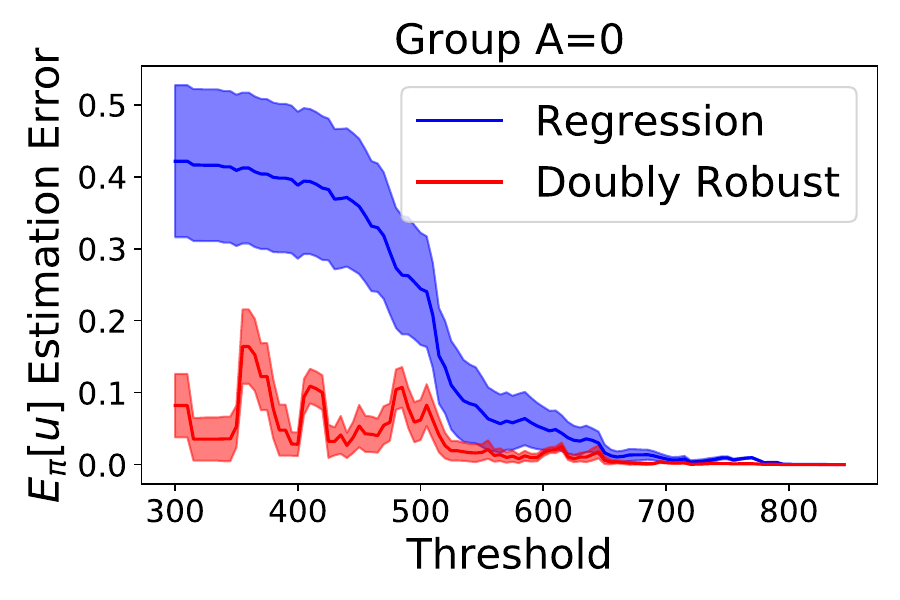}
    }
  \end{subfigure}
  \hfill
  \begin{subfigure}[t]{0.23\textwidth}
    \noindent\resizebox{\textwidth}{!}{
      \includegraphics[]{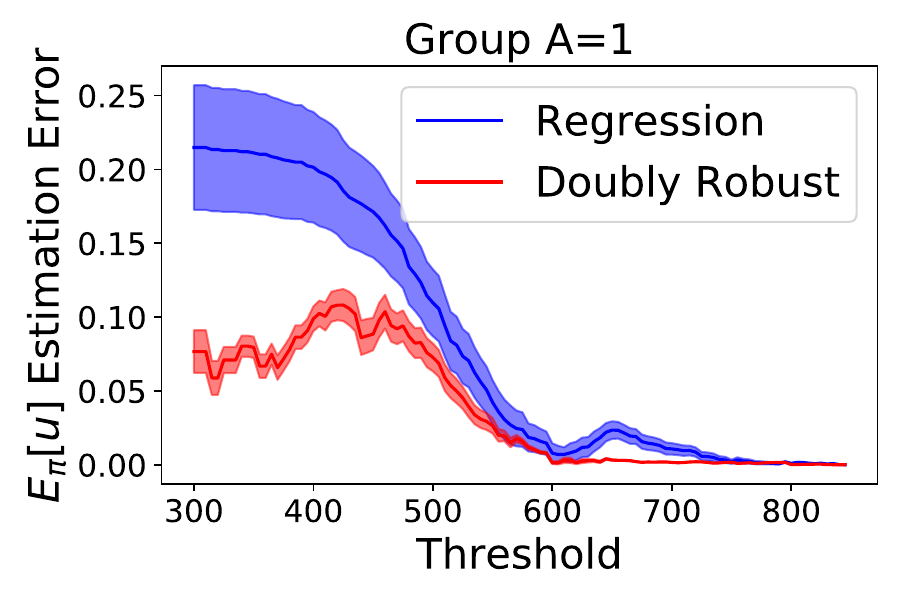}
    }
  \end{subfigure}
  \caption{
  Comparing error of $\mathcal{E}_{Reg}$ and $\mathcal{E}_{DR}$ (regression and doubly robust) for off-policy estimation of $\mathds{E}_{\pi}[u]$ from observational data.
  The x-axis represents single threshold policies.
  Estimation for $\Delta$ yields similar results, since both variables are linear in $Y$.
}\label{fig:estimation-reg-vs-dr}
\end{figure}

We generate observational data from the SCM in Figure \ref{fig:scm-liu}, under a \MaxProf \ threshold policy.
We then consider a new policy $\pi_\tau$ with per-group thresholds $\{\tau_j\}$ as its two parameters.
We compute the estimators $\mathcal{E}_{Reg}(\pi_\tau)$ and $\mathcal{E}_{DR}(\pi_\tau)$ for varying values of these thresholds.
Figure \ref{fig:estimation-reg-vs-dr} shows that the causally motivated estimator $\mathcal{E}_{DR}$ achieves lower off-policy estimation error on both sensitive groups, across the threshold range.
Note the high estimation error of the baseline $\mathcal{E}_{Reg}$ for low values of $\tau$.
This is because the historical policy typically does not award loans to applicants with low scores, meaning there are fewer data available for the regression.

Ultimately, the goal of estimating these quantities is to improve policy learning.
We can formulate an objective which trades off between utility and an equal opportunity term {\small ${\delta_{EqOpp} = | P(T = 1 | Y = 1, A = 0) -  P(T = 1 | Y = 1, A = 1)|}$}.
The overall objective is $\mathcal{V}_{\pi} = \mathcal{U} - \lambda \delta_{EqOpp}$. We hope to maximize this, with some hyperparameter $\lambda \in \mathds{R}$ governing the tradeoff.
We note that estimating $\delta_{EqOpp}$ itself presents a challenging causal problem, since $Y$ is frequently unobserved.
See Appendix \ref{app:ope-details} for details on this estimation problem and the rest of this experiment.

\begin{figure}[ht!]
  \centering
      \includegraphics[scale=0.33]{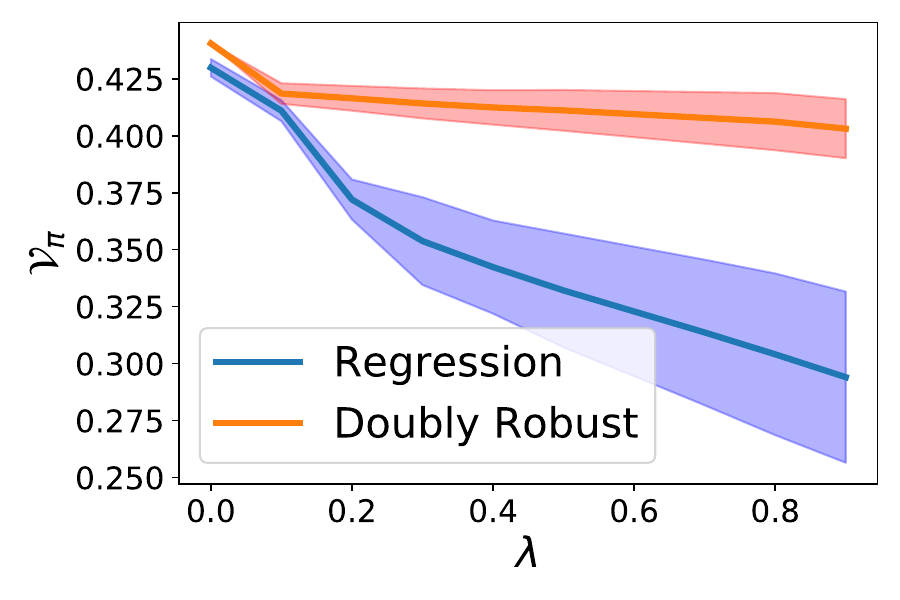}
  \caption{Test set value of a fairness-utility objective using the two off-policy estimators.
  Hyperparameter $\lambda$ governs the tradeoff.
  Higher values of the objective $\mathcal{V}_{\pi}$ are better.
  }\label{fig:policy-learning-reg-vs-dr}
\end{figure}

Using the estimators presented above, we can construct an off-policy estimate of $\mathcal{V}_{\pi}$.
We search over the space of two-threshold policies (one threshold per group) to find the policy with the highest off-policy estimate of the objective on a validation set.
We then calculate the true value 
of $\mathcal{V}_{\pi}$ on a held-out test set, using the SCM simulator (as visualized in Figure \ref{fig:scm-liu} and specified fully in Appendix \ref{app:scm-liu}) to generate the true potential outcomes.
The estimator $\mathcal{E}_{DR}$ that more fully incorporates causal reasoning in the parameter estimation finds a better objective value, ultimately yielding an improved policy (see Fig. \ref{fig:policy-learning-reg-vs-dr}).
We emphasize that this improvement requires assumptions about causal structures, but not precise knowledge of the system dynamics.

\section{Extensions in Lending via Intervention}\label{sec:individual-dynamics}
We now investigate the setting where both causal structure \emph{and} dynamics are known (returning to the assumptions made by \citet{liu2018delayed}), and emphasize the role of \emph{interventions} in building expressive simulators for dynamical fairness settings.
Thus we carry out ``on policy'' evaluations that sample from the SCM directly.
SCMs enable clearer explication of underlying causal assumptions.
This means the framework is flexible: novel policy interventions extend our model by modifying existing assumptions, or testing our reliance on the assumptions we have already made.
We give two such examples, measuring:
(a) the interaction of the lender with other agencies; and (b) the sensitivity of long-term outcomes to the lender's modeling assumptions.

\subsection{Multi-actor Experiments}\label{sec:scm-liu-extensions-multi-actor}
\paragraph{Intervention by credit bureau}
\citet{liu2018delayed} conduct experiments based on statistics of FICO credit scores 
assigned by the credit bureau TransUnion \citep{reserve2007report}.
We note that these credit score decisions themselves constitute a policy; and
moreover, the language of interventions in the SCM framework allows us to
characterize decisions made by the \emph{credit bureau} (rather than the bank)
using the same fairness and profit metrics as before.\footnote{Note that recent changes by the credit scoring bureau 
Fair Isaac Corp. (\url{https://www.wsj.com/articles/fico-changes-could-lower-your-credit-score-11579780800})
can be characterized as such an intervention.}
The credit bureau enters the SCM by reinterpreting
$X_i$ as \emph{features} related to creditworthiness of an individual, then
introducing $\hat X_i = f_{\hat X}(X_i)$ as a \emph{score} that is
deterministically computed by the agency from the features (See Fig. \ref{fig:scm-liu-extras}).
When $f_{\hat X}$ is the identity function, we recover the original model.
Policy evaluation under double intervention $\mathcal{M}^{\text{do}(f_T
\rightarrow \hat f_T, f_{\hat X} \rightarrow \hat f_{\hat X})}$ captures the
sensitivity of the bank's decisions to the decisions of the credit bureau (and
vice versa).
\begin{figure}[t!]
  \begin{subfigure}[b]{.23\textwidth}
    \begin{center}
        \includegraphics[scale=0.25]{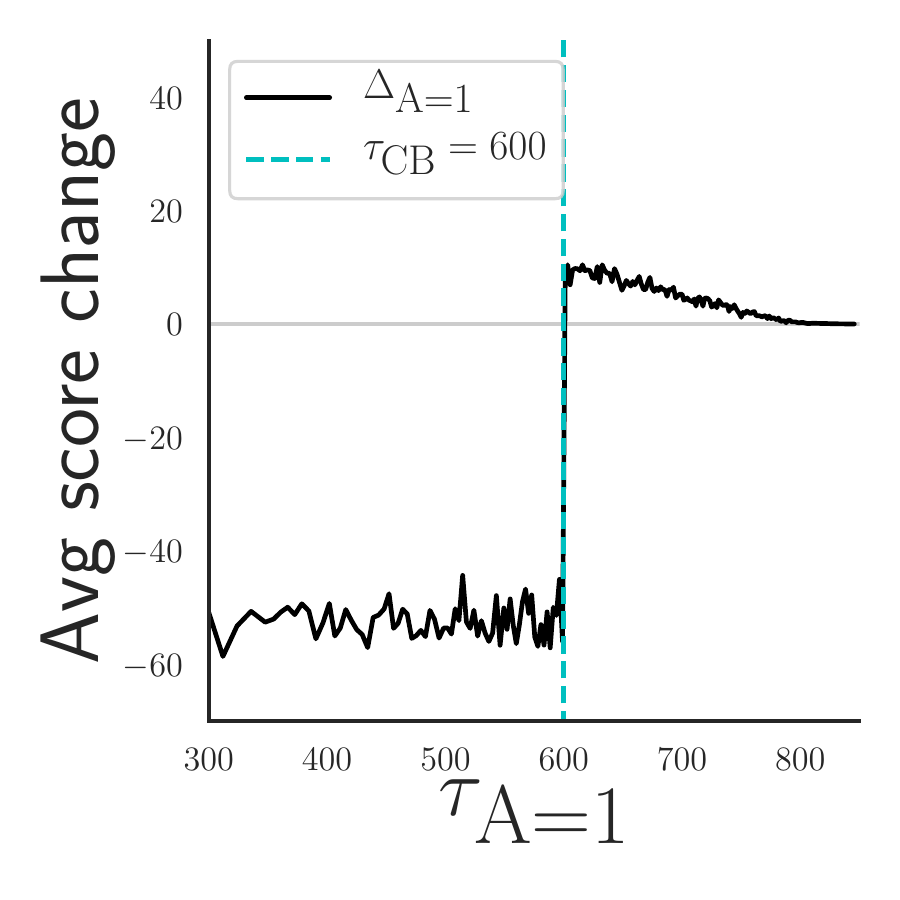}
        \caption{
          Score change, min. group.
            }\label{fig:credit_score_intervention_a}
    \end{center}
  \end{subfigure}
  \hfill
  \begin{subfigure}[b]{.23\textwidth}
    \begin{center}
        \includegraphics[scale=0.25]{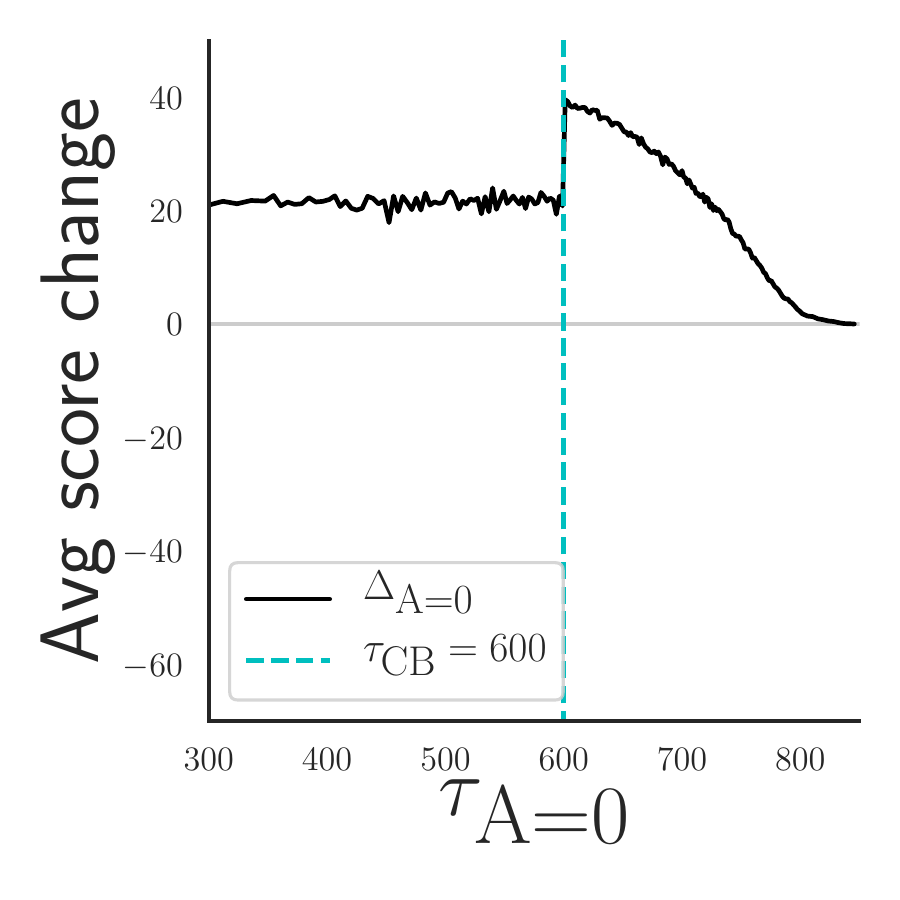}
        \caption{
          Score change, maj. group.
        }\label{fig:credit_score_intervention_b}
    \end{center}
  \end{subfigure}
  \vskip\baselineskip
  \begin{subfigure}[b]{.23\textwidth}
    \begin{center}
        \includegraphics[scale=0.25]{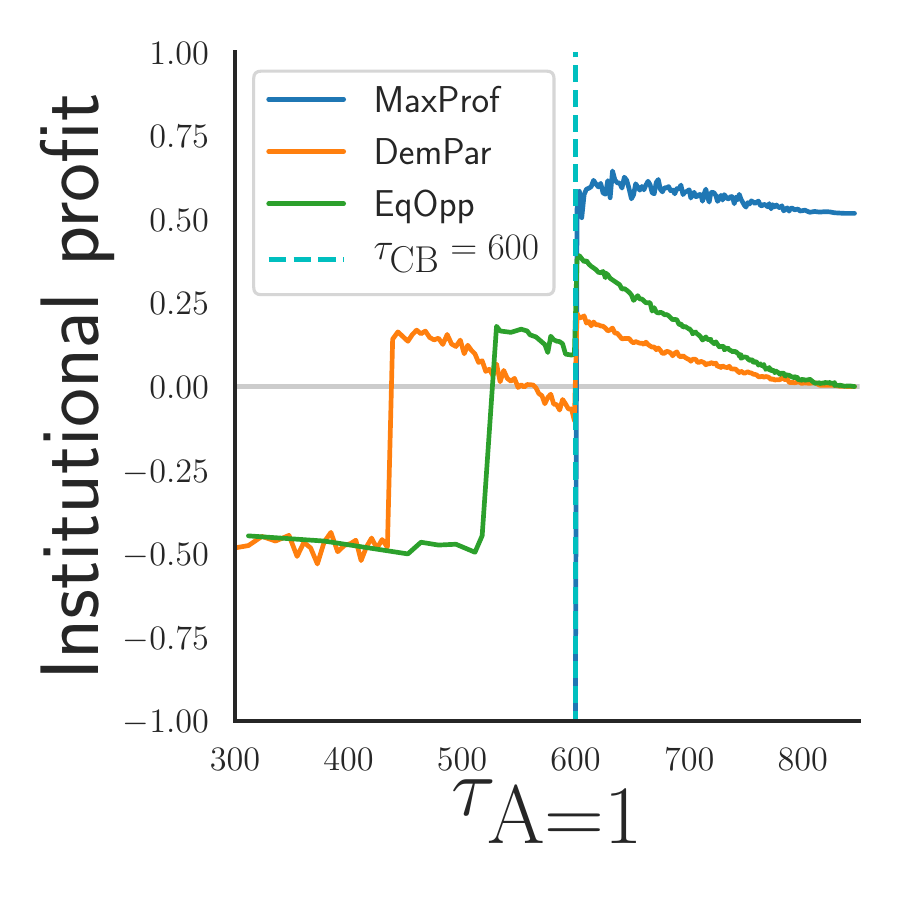}
        \caption{
         Profit as fn. of min. thresh.
        }\label{fig:credit_score_intervention_c}
    \end{center}
  \end{subfigure}
  \hfill
  \begin{subfigure}[b]{.23\textwidth}
    \begin{center}
        \includegraphics[scale=0.25]{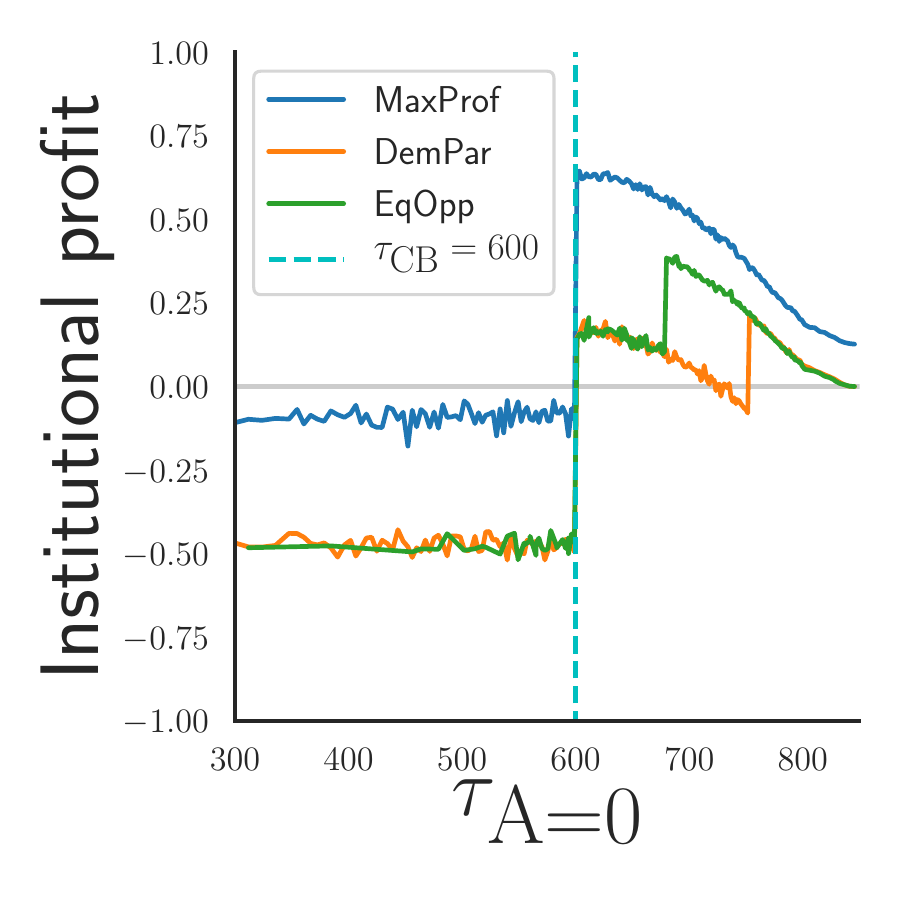}
        \caption{
         Profit as fn. of maj. thresh.
        }\label{fig:credit_score_intervention_d}
    \end{center}
  \end{subfigure}
  \caption{
    Policy evaluation under credit bureau intervention
    $\hat f_{\hat X}(X) = \text{min}(X, \tau_{CB})$ with $\tau_{CB}=600$.
    Group score change---formally $\E_{p^{do(f_{\hat X} \rightarrow \hat f_{\hat
    X}, f_{T} \rightarrow \hat f_T)}} [\Delta_j]$---and 
    institutional profits---$\E_{p^{do(f_{\hat X}
    \rightarrow \hat f_{\hat X}, f_{T} \rightarrow \hat f_T )}}
    [\mathcal{U}]$---are shown as functions of the two group thresholds
    $\{\tau_j\}$ under several fairness constraints.
  }\label{fig:case-study-first-results}
\end{figure}
\paragraph{Results}
Figure~\ref{fig:case-study-first-results} shows the effect on the average
utility $\E[\mathcal{U}]$ and average per-group score change $\E[\Delta_j]$ of a
simple policy intervention by the credit bureau.
The intervention involves the bureau setting the minimum score to $600$ for all
applicants via the structural equation ${\hat f_{\hat X}(X) = \text{min}(X,
600)}$.
This intervention is unlikely in the real world because it contradicts the
profit incentives of the bureau, which encourage well-calibrated scores.
Nevertheless, it coarsely captures a potential scenario where an actor besides
the bank seeks to encourage fair outcomes in a group-blind way, since under the
new scoring policy minority applicants are more likely to receive loans.
However, we see in Figure~\ref{fig:credit_score_intervention_a} that the average
group outcome for protected applicants ($A=1$) worsens when the bank's group threshold
$\tau_{A=1}$ is below 600, since in this case its policy offers loans to
individuals who have good scores on paper but are unlikely to repay the loans.
Interestingly, the expected profit
(Figure~\ref{fig:credit_score_intervention_b}) under credit bureau intervention
differs depending on the fairness criteria of the bank.
This is because each fairness criteria differently constrains the relationship
between the two thresholds $\{\tau_{A=0}, \tau_{A=1}\}$ (the protected group is $A=1$), so the choice of
fairness criteria implicitly sets how many applicants with boosted scores
($X < 600$, thus $\hat X = 600$) are selected for loans.
\DP \ is more sensitive to the credit bureau intervention than \EqOpp; it obeys a stricter fairness constraint and offers more loans to 
applicants with boosted scores (who are are unlikely to repay, and
disproportionately belong to the minority group).

\subsection{Sensitivity Analysis of Long-term Outcomes}\label{sec:scm-liu-extensions-multi-step}
Sensitivity analysis \citep{rosenbaum2014sensitivity,saltelli2008global}---the task of measuring how sensitive a system's output is to its various assumptions---is critical when engaging in a complex modeling task.
Since causal language makes structural modeling assumptions explicit, it is a natural match for sensitivity analysis.
Questions of robustness are particularly important in long-term, dynamic modeling, since small assumptions errors can have large effects downstream when propagated over time.
In this section, we show how to conduct a long-term sensitivity analysis of the \citet{liu2018delayed} model with SCMs, probing how sensitive proposed policies may be to underlying causal assumptions.
We cast the sensitivity analysis as an on-policy evaluation under an intervention that accounts for
model mismatch.\footnote{
``Mismatch'' refers here to structural equations with misspecified functional forms, not incorrect assumptions of causal structure.
}
\begin{figure}[t!]
\centering
\noindent\resizebox{.5\textwidth}{!}{
  \newcommand{\ssi}{4}  %
\newcommand{\exu}{.3cm}  %
\begin{tikzpicture}[scale=0.85, >=stealth]
  \tikzstyle{lat-exo}=[circle, inner sep=1pt, minimum size = 6.5mm, thick, draw=black!80, node distance = 20mm, scale=0.75, font=\large]
  \tikzstyle{obs-exo}=[circle, fill=black!20, inner sep=1pt, minimum size = 6.5mm, thick, draw=black!80, node distance = 20mm, scale=0.75]
  \tikzstyle{lat-box}=[rectangle, inner sep=1pt, minimum size = 6mm, draw=black!80, thick, node distance = 20mm, scale=0.75, font=\huge]
  \tikzstyle{obs-box}=[rectangle, fill=black!20, inner sep=1pt, minimum size = 6mm, thick, draw=black!80, node distance = 20mm, scale=0.75, font=\huge]
\tikzstyle{inv-box}=[rectangle, fill=black!20, inner sep=1pt, minimum size = 6mm, thick, draw=black!80,densely dotted, node distance = 20mm, scale=0.75]
  \tikzstyle{directed}=[-latex, thick, shorten >=0.5 pt, shorten <=1 pt]
  \tikzstyle{interven}=[directed, densely dotted]
  \tikzstyle{empty}=[]

\node[obs-box] (a)                              at (0,0)         {$A$};
\node[lat-exo]  (ua)              [above left=\exu of a]     {$U_{A}$};
\node[obs-box]  (x0)                            at (1,3)       {$X^0$};
\node[lat-exo] (ux0)                  [above=\exu of x0]   {$U_{X^0}$};
\node[obs-box]  (y0)                            at (2,1)       {$Y^0$};
\node[lat-exo] (uy0)                   [left=\exu of y0]   {$U_{Y^0}$};
\node[obs-box]  (t0)                            at (4,1)       {$T^0$};
\node[lat-exo] (ut0)                   [left=\exu of t0]   {$U_{T^0}$};
\node[obs-box]   (u0)                           at (4,4)       {$u^0$};
\node[obs-box]  (x1)                       at (1+\ssi,3)       {$X^1$};
\node[lat-exo] (ux1)            [above right=\exu of x1]   {$U_{X^1}$};
\node[obs-box]  (y1)                       at (2+\ssi,1)       {$Y^1$};
\node[lat-exo] (uy1)                   [left=\exu of y1]   {$U_{Y^1}$};
\node[obs-box]  (t1)                       at (4+\ssi,1)       {$T^1$};
\node[lat-exo] (ut1)                   [left=\exu of t1]   {$U_{T^1}$};
\node[obs-box]  (u1)                       at (4+\ssi,4)       {$u^1$};
\node[obs-box]  (x2)                     at (1+2*\ssi,3)       {$X^2$};
\node[lat-exo] (ux2)            [above right=\exu of x2]   {$U_{X^2}$};
\node[obs-box]  (y2)                     at (2+2*\ssi,1)       {$Y^2$};
\node[lat-exo] (uy2)                   [left=\exu of y2]   {$U_{Y^2}$};
\node[obs-box]  (t2)                     at (4+2*\ssi,1)       {$T^2$};
\node[lat-exo] (ut2)                   [left=\exu of t2]   {$U_{T^2}$};
\node[obs-box]  (u2)                     at (4+2*\ssi,4)       {$u^2$};
\node[obs-box]   (U)                     at (3*\ssi,5) {$\mathcal{U}$};
\node[obs-box]  (Dj)                   at (3*\ssi-3,6.5)    {$\Delta$};
\node[empty]     (e)                [right=\exu of Dj]              {};
\path
         (ua) edge [directed]  (a)
        (ux0) edge [directed] (x0)
         (x0) edge [directed] (t0)
          (a) edge [directed] (y0)
          (a) edge [directed] (x0)
          (a) edge [directed, bend right=10] (t0)
        (ut0) edge [directed] (t0)
         (x0) edge [directed] (y0)
        (uy0) edge [directed] (y0)
         (y0) edge [directed] (u0)
         (t0) edge [directed] (u0)
        (x0)  edge [directed, bend right=30] (x1)
        (y0)  edge [directed, bend right=10] (x1)
        (t0)  edge [directed] (x1)
        (ux1) edge [directed] (x1)
         (x1) edge [directed] (t1)
          (a) edge [directed, bend right=15] (y1)
          (a) edge [directed, bend right=20] (t1)
        (ut1) edge [directed] (t1)
         (x1) edge [directed] (y1)
        (uy1) edge [directed] (y1)
         (y1) edge [directed] (u1)
         (t1) edge [directed] (u1)
        (x1)  edge [directed, bend right=30] (x2)
        (y1)  edge [directed, bend right=10] (x2)
        (t1)  edge [directed] (x2)
        (ux2) edge [directed] (x2)
         (x2) edge [directed] (t2)
          (a) edge [directed, bend right=25] (y2)
          (a) edge [directed, bend right=30] (t2)
        (ut2) edge [directed] (t2)
         (x2) edge [directed] (y2)
        (uy2) edge [directed] (y2)
         (y2) edge [directed] (u2)
         (t2) edge [directed] (u2)
        (x0) edge  [directed, bend left=10] (Dj)
        (x2) edge [directed, bend right=20] (Dj)
        (u0) edge  [directed, bend left=10]  (U)
        (u1) edge  [directed, bend left=05]  (U)
        (u2) edge [directed]                 (U)
        ;
\node[empty]   (ea)              [below=\exu of a] {};
\plate {} {
    (ua)(a)(ea)
    (x0)(ux0)(t0)(ut0.north west)(y0)(uy0)(u0)
    (x1)(ux1)(t1)(ut1.north west)(y1)(uy1)(u1)
    (x2)(ux2)(t2)(ut2.north west)(y2)(uy2)(u2)
  } {$N$} ;
\plate {} {(Dj)(e)} {$|\mathcal{A}|$} ;
\end{tikzpicture}
}
\caption{
  Phrasing the model from \citet{liu2018delayed} as an SCM enables a multi-step
  extension for measuring long-term impacts, e.g., in the two-step version shown
  here.
  }\label{fig:scm-liu-multi-step}
\end{figure}
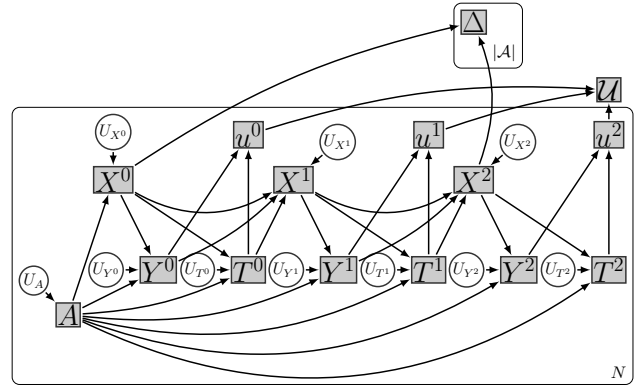

\begin{figure*}[ht!]
\centering
  \begin{subfigure}[b]{.45\textwidth}
    \begin{center}
        \includegraphics[width=.8\textwidth]{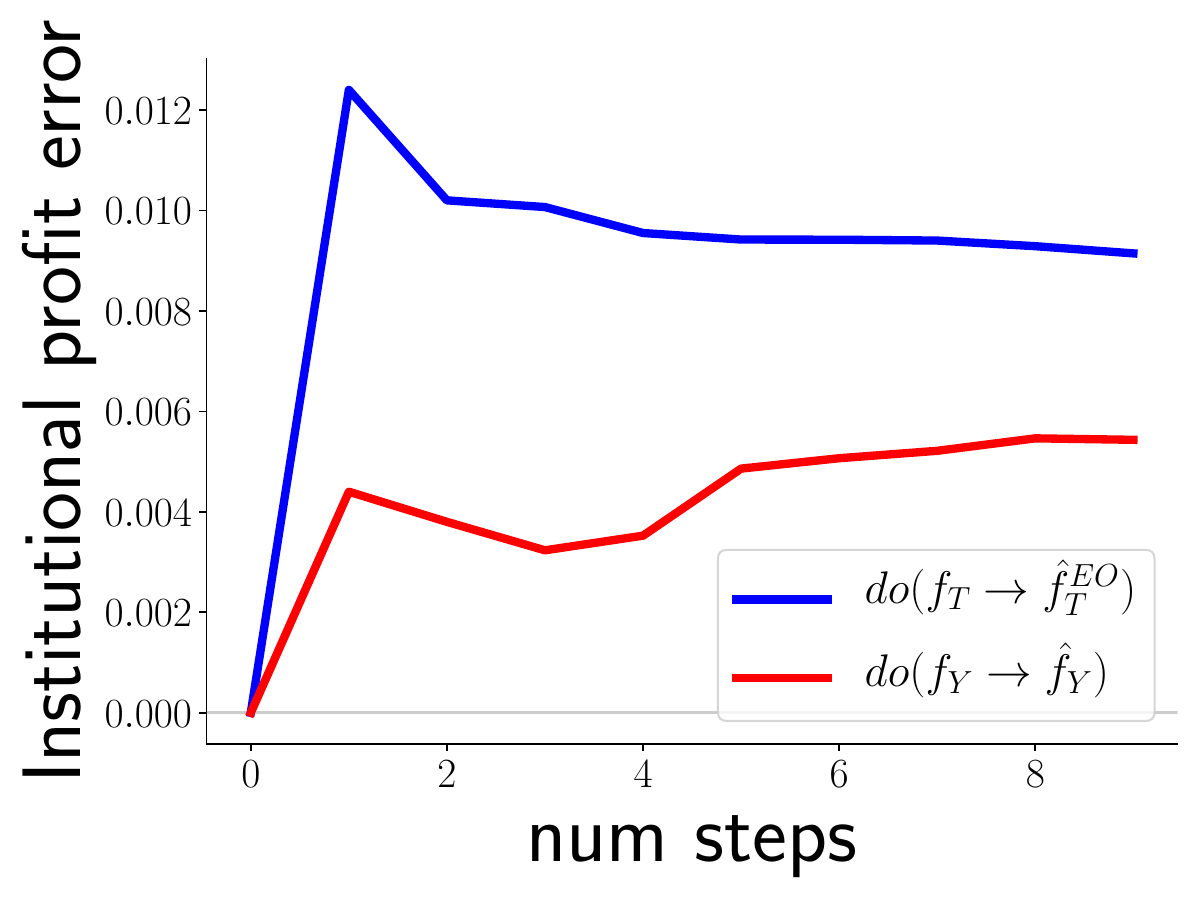}
    \end{center}
  \end{subfigure}
  \begin{subfigure}[b]{.45\textwidth}
    \begin{center}
        \includegraphics[width=.8\textwidth]{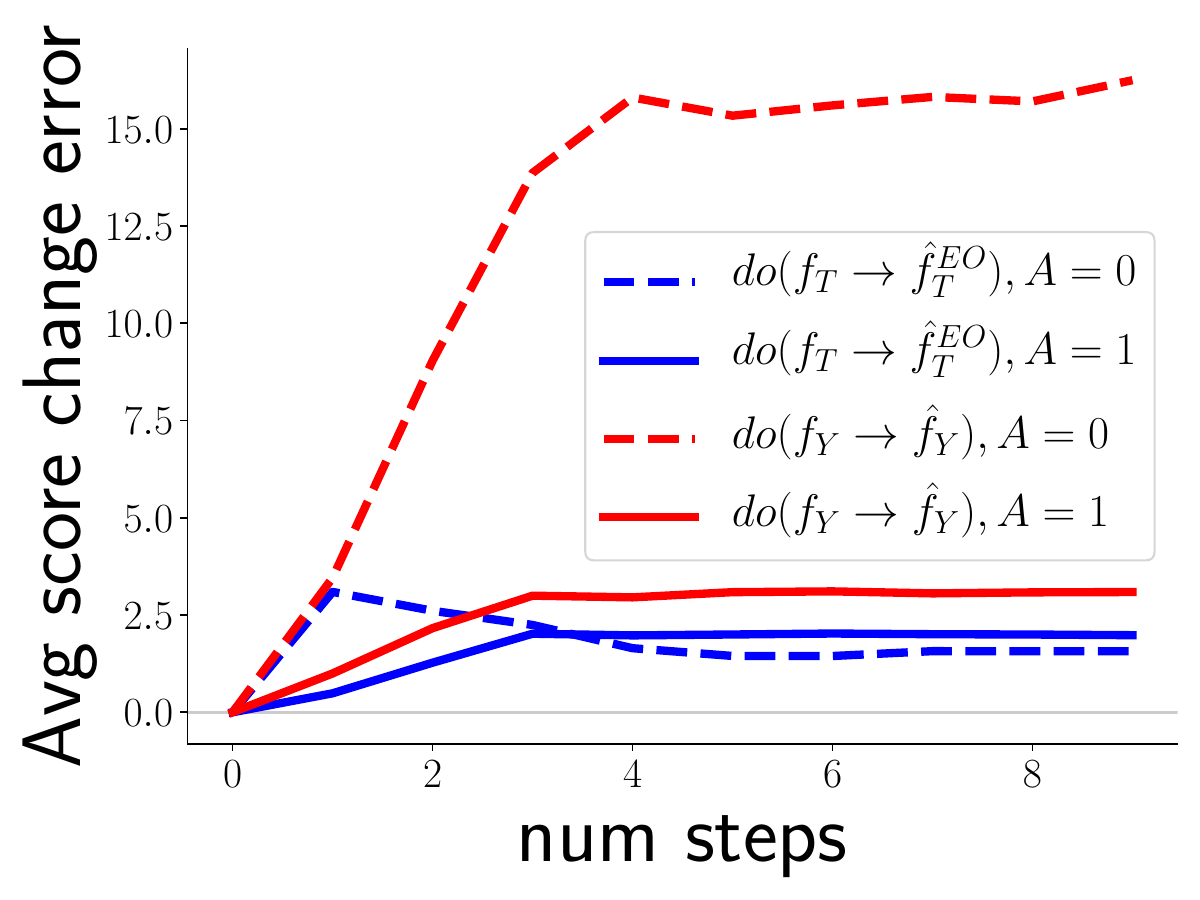}
    \end{center}
  \end{subfigure}
  \caption{
    Evaluating multi-step policy robustness to distribution shift for various
    choice of intervention distribution $q$.
    Sensitivity of institutional utility---formally $|\E_q[\mathcal{U}] -
    \E[\mathcal{U}]|$---and sensitivity of group avg. score change---formally
    $|\E_q[\Delta_j] - \E[\Delta_j]|$---are shown as a function of steps.
    Expected profit is relatively robust to both interventions, whereas the
    expected per-group score changes are relatively more sensitive to these
    interventions.
  }\label{fig:case-study-second-results}
\end{figure*}

\paragraph{Long-term impacts}
Given a policy whose one-step effect is purportedly fair, what can we say about
its longer-term impacts?
The modularity of the SCM formulation allows us to easily estimate these
effects.
For example, the structural equation $\tilde X~=~f_{\tilde X}(X, Y, T)$
can be modified to the recursive update $X^{t+1} = f_{X^{t > 0}}(X^t, Y^t,
T^t)$.
Note that $X^t$ (which does not directly depend on $A$) is only computed in
this way for steps $t > 0$, since the original scores $X^0$ are sampled from
$p(X^0|A)$.
On the other hand, since $T^t$ and $Y^t$ still depend on $A\ \forall\ t$, we see that group membership does indeed have a long-term influence on the outcomes and score trajectories for individuals.

\paragraph{Results}
We conduct our sensitivity analyses as simulations of policy interventions under varying underlying model assumptions.
We analyze the sensitivity of the \EqOpp \  policy to two forms of model mismatch.
In the first, $\text{do}(f_T \rightarrow \hat f_T^{EO})$ recomputes the per-group thresholds
under the \EqOpp \  constraint, but using \emph{incorrect} statistics from the credit bureau.
In particular, the marginal $p(Y|X)$ was used for both group's repayment probabilities rather than the correct $p(Y|X,A)$.
The second intervention $\text{do}(f_Y \rightarrow \hat f_Y)$ is more severe, as $p(Y|X)$ is used to \emph{sample} potential outcomes $Y$ rather than just set the thresholds within $f_T$.
We measure error under each intervention relative to the ``ground truth'' baseline where the correct potential outcome distributions are used to set thresholds and sample data.
We measure how these errors compound over time (Figure \ref{fig:case-study-second-results}).
We observe the institutional profits are surprisingly robust to both forms of intervention, while the per-group outcomes are more sensitive to these interventions, especially to
$\text{do}(f_Y \rightarrow \hat f_Y)$.
These results indicate that our policies are particularly sensitive to assumptions around outcome prediction for the sensitive groups.


\section{Discussion}\label{sec:discussion}
In this paper, we discuss causal modelling as a unifying framework for the literature on fairness in dynamical systems.
We demonstrate that in the realistic situation where the dynamics parameters of these systems are not known, causal models are helpful for estimating these parameters, and evaluating and learning policies in an off-policy manner from historical data.
Additionally, we show how a causal model can be used as a simulator when the parameters \textit{are} known, and how the modularity of the SCM framework is helpful for both expressing natural extensions existing work from the literature, and running long-term sensitivity analyses of policy decisions.

Since a causal DAG can be thought of as an expressive simulator, standard tools for optimization/learning in computation graphs
\citep{schulman2015gradient} can be brought to bear
in order to \emph{learn} policies that capture optimal rewards 
\emph{across many interventional settings}.
Using gradient estimators to learn policies in this setting holds promise in scaling to high dimensional datasets, which we leave for future work.

\section*{Acknowledgements}
We thank Kuan-Chieh Wang, Eleni Triantafillou, Will Grathwohl, James Lucas, Robert Adragna, and John Miller for providing feedback, and the reviewers for their helpful suggestions.

Resources used in preparing this research were provided, in part, by the Province of Ontario, the Government of Canada through CIFAR, and companies sponsoring the Vector Institute \url{www.vectorinstitute.ai/#partners}.

\bibliographystyle{icml2020}
\bibliography{refs}
\clearpage
\onecolumn

\appendix
\section{Experimental Details for Off-Policy Evaluation and Selection} \label{app:ope-details}

Here, we discuss details on the setup for the off-policy evaluation experiment in 
Sec. \ref{sec:off-policy-eval}.

\subsection{Data Generation}

We generate data from the \citet{liu2018delayed} model, described in full in Appendix \ref{app:scm-liu}.
We use $(c_+, c_-) = (75, -150)$ and $(u_+, u_-) = (1, -4)$.
We use a single threshold policy of $\tau_j = 620 \forall \ j$.
We generate 13 data sets of $10,000$ examples each, using 11 for training (to get confidence intervals), 1 for validation, and 1 for test.

In order to use re-weighting estimators, we must have \textit{overlap} i.e. each point $(X, A)$ must have a non-zero probability of receiving each treatment in the observational data.
Since a threshold policy does not satisfy this, we flipped the treatment chosen by the threshold policy with a probability of 0.1.

\subsection{Treatment and Outcome Models}

We use L2-regularized logistic regression for both the treatment and the outcome model using the \texttt{liblinear} default solver in sklearn.
We train a treatment and outcome model on each of the 11 training sets, and use these to construct our confidence intervals.

\subsection{Estimation of Equal Opportunity Distance}
We define the equal opportunity metric $\delta_{EqOpp}$ as
\begin{align}
    \delta_{EqOpp} &= | P(T = 1 | Y = 1, A = 0)  -  P(T = 1 | Y = 1, A = 1) |.
\end{align}
The key unit in this expression is $P(T = 1 | Y = 1)$ (removing $A = a$ from the right side for clarity).
This is non-trivial to estimate, since $Y$ is unobserved for many cases.

We take the following approach.
First, using Bayes rule, we have
\begin{equation}
    P(T = 1 | Y = 1) = \frac{ P(Y = 1 | T = 1) P(T = 1) }{P(Y = 1)}.
\end{equation}{}
$P(T = 1)$ is easy to estimate from observational data.
$P(Y = 1 | T = 1)$ is the off-policy estimation question --- we use either $\mathcal{E}_{Reg}$ or $\mathcal{E}_{DR}$ to estimate this.
We estimate $P(Y = 1)$ using off-policy estimation as well, noting that $P(Y = 1) = P(Y = 1 | \Tilde{T} = 1)$, if $\Tilde{T} \perp Y$.
Therefore, we can obtain an estimate for the marginal distribution of $Y$ by doing off-policy estimation for random policies $\Tilde{T}$ (again, using either $\mathcal{E}_{Reg}$ or $\mathcal{E}_{DR}$).
We choose 10 random Bernoulli policies to obtain 10 estimates of $P(Y = 1)$ and average them.

\subsection{Threshold Search}

In both the estimation (Fig. \ref{fig:estimation-reg-vs-dr}) and selection (Fig. \ref{fig:policy-learning-reg-vs-dr}) experiments, we consider all thresholds\footnote{
300 and 850 are the minimum and maximum credit scores in the dataset
} $\tau \in [300, 850)$ in increments of $5$.
To choose our best thresholds in the selection experiment, we consider all pairs of group-specific thresholds $(\tau_0, \tau_1)$, and estimate the value of $\mathcal{V_{\pi}}$ for the policy associated with those thresholds.
We find the optimal value on the validation set, and test them to obtain a final value on the test set,
Since we do not require overlap to hold in the target policy, we consider hard threshold policies (we do not flip any predictions post-hoc, as we do in the observational data).
In the selection experiment, we test $\lambda$ in increments of 0.1 from 0 to 0.9.

\section{\citet{liu2018delayed} SCM Details}\label{app:scm-liu}
As briefly discussed above, \citet{liu2018delayed} propose a one-step
feedback model for a decision-making setting then analyze several candidate
policies---denoted by the structural equation $f_T$ in our analysis---by
simulating one step of dynamics to compute the institution's profit and group
outcomes for each policy.
Figure~\ref{fig:scm-liu} shows our SCM formulation of this dynamics model.
Here we provide expressions for the specific structural equations used.
Throughout, we make the assumption that our model and its associated counterfactuals are representative of the observed data --- this is termed as the \textit{consistency} assumption, and is described by \citet{pearl2010consistency} as 
\begin{equation}
    P(Y_x = y | Z = z, X = x) = P(Y = y | Z = z, X = x)
\end{equation}
for all $x, y, z$, where $Y_x$ is the counterfactual potential outcome for $Y$ under the treatment $x$.

To sample over $p(X, A)$ we start with Bernoulli sampling of $A$, parameterized
  SCM-style like
\begin{align}
  U_{A_i} & \sim \text{Bernoulli}(U_{A_i}|\theta);\ \ \ A_i = f_A(U_{A_i}) \triangleq U_{A_i}
\end{align}
where $\theta \in [0, 1]$ is the proportion of the $A=1$ group.

We then sample scores by the inverse CDF trick\footnote{
This standard trick is used for sampling from distributions with know densities.
Recalling that ${\text{CDF}_p: \mathcal{X} \rightarrow [0, 1]}$ is a monotonic
(invertible) function representing ${\text{CDF}_p(X') = \int_{-\infty}^{X'}dX
p(X < X')}$.
Then to sample {$X' \sim p$} we first sample ${U \sim \text{Uniform}(U|[0,1])}$
then compute ${X' = \text{CDF}_p^{-1}(U)}$.
}.
Given an inverse cumulative distribution function $\text{CDF}^{-1}_j$ for each
group $j \in \{0, 1\}$, we can write
\begin{align}
  U_{X_i} & \sim \text{Uniform}(U_{X_i}|[0, 1]) \\
  X_i & = f_X(U_{X_i}, A_i) \triangleq \text{CDF}^{-1}_{A_j}(U_{X_i}) \label{eq:init-score}
\end{align}

\citet{liu2018delayed} discuss implementing threshold policies for each group
  $j \in \{0, 1\}$, which are parameterized by thresholds $c_j$ and tie-breaking Bernoulli probabilities $\gamma$ (for simplicity of exposition we assume the tie-breaking probability is shared across groups).
The original expression was
\begin{align}
  \mathbb{P}(T=1|X, A=j) &= \
  \begin{cases}
    1 & X > c_j \\
    \gamma & X = c_j \\
    0 & X < c_j.
  \end{cases}
\end{align}

Then, after denoting by $\ind{\cdot}$ the indicator function, we can rephrase this distribution in terms of a structural equation governing treatment:
\begin{align}
  U_{T_i} & \sim \text{Bernoulli}(U_{T_i}|\gamma) \\
  T_i & = f_T(U_{T_i}, X_i, A_i) \nonumber \\
      & \triangleq 1^{\ind{X_i > c_{A_i}}} \cdot U_{T_i}^{\ind{X_i = c_{A_i}}} \cdot 0^{\ind{X_i < c_{A_i}}}. \label{eq:treatment}
\end{align}

A policy $f_T$ (which itself may or may not satisfy some fairness criteria) is evaluated in terms of whether loans were given to creditworthy individuals, and in terms of whether each demographic group successfully repaid any allocated loans on average.
To capture the notion of \emph{creditworthiness}, we introduce a
  \emph{potential outcome} $Y$ (repayment if the loan were given) for each
  individual, which is drawn\footnote{
    The authors denoted by $\boldsymbol{\rho}(x)$ the probability of potential
    success at score $X$.  Various quantities were then computed, e.g.,
    $\boldsymbol{u}(x) = u_+ \boldsymbol{\rho}(x) + u_- (1 -
    \boldsymbol{\rho}(x))$.  We observe that this is equivalent to
    marginalizing over potential outcomes $\boldsymbol{u}(x) =
    \E_{p(Y|X)}\left[ u_+ Y + u_- (1 - Y)\right]$; in our simulations we
    compute such expectations via Monte Carlo sampling with values of $Y$
    explicitly sampled.
  } from $p(Y|X, A)$\footnote{
    The authors use $\boldsymbol{\rho}(X)=p(Y|X)$ in their analysis
    (suggesting that potential outcome is independent of group membership
    conditioned on score) but $\boldsymbol{\rho}(X,A)=p(Y|X,A)$ in the
    code, i.e. the potential outcome depends differently on score for each
    group.
    The SCM as expressed in Figure~\ref{fig:scm-liu} represents the codebase
    version.
  }.
By convention $T=1$ as the ``positive'' treatment (e.g., got loan) and $Y=1$ as the ``positive'' outcome (e.g., would have repaid loan if given)
Note that $Y$ is independent of $T$ given $X$, meaning $Y$ is really an indicator of \emph{potential} success.
Formally, the potential outcome $Y$ is distributed as
$Y_i \sim \text{Bernoulli}(Y_i|\boldsymbol{\rho}(X_i, A_i))$
for some function $\boldsymbol{\rho}: X \times A \rightarrow[0, 1]$.
We reparameterize this as a structural equation using the Gumbel-max trick\footnote{
This trick reparameterizes a Categorical or Bernoulli sample as a deterministic transformation of a Uniform sample. 
See \citet{oberst2019counterfactual} for discussion of how to perform counterfactual inference for SCMs with Categorical
random variables.
}
  \citep{gumbel1954some, maddison2014sampling}:
\begin{align}
  U_{Y_i} & \sim \text{Uniform}(U_{Y_i}|[0, 1]) \\
  Y_i & = f_{Y}(U_Y, X_i, A_i) \triangleq \mathbb{1} \left( \log \frac{\boldsymbol{\rho}(X_i, A_i)}{1 - \boldsymbol{\rho}(X_i, A_i)} + \log \frac{U_Y}{1 - U_Y} > 0 \right). \label{eq:potential-outcome}
\end{align}

The institutional utility $u_i$ and the updated individual score $\tilde X_i$ are deterministic functions of the outcome $Y_i$ and the treatment $T_i$, and
  the original score $X_i$:
\begin{align}
  u_i & = f_{u}(Y_i, T_i) \triangleq
  \begin{cases}
    u_+^{\ind{Y_i} = 1} \cdot u_-^{\ind{Y_i} = 0} & \text{if} \ T_i = 1 \\
    0                                         & \text{else}
  \end{cases}, \\
  \tilde X_i & = f_{\tilde X}(X_i, Y_i, T_i) \triangleq
  \begin{cases}
    X_i + c_+^{\ind{Y_i} = 1} \cdot c_-^{\ind{Y_i} = 0} & \text{if} \ T_i = 1 \\
    X_i                                                       & \text{else}
  \end{cases}.
\end{align}
As mentioned in Section~\ref{sec:off-policy-eval}, $\{u_+, u_-, c_+, c_-\}$ are fixed parameters that encode
  expected gain/loss in utility/score based on payment/default of loan.

There are two \emph{global} quantities of interest.
Firstly, the institution cares about its overall utility at the current step  (ignoring all aspects of the future), expressed as
\begin{align}
  \mathcal{U} = f_{\mathcal{U}}(u_{1 \ldots N}) \triangleq \frac{1}{N} \sum_{i=1}^N u_i.
\end{align}
Secondly, to understand the societal impact of the lending policy, we measure the average per-group score change induced by the policy, expressed for group $A=j$ as
\begin{align}
  \Delta_j &= f_{\Delta_j}(X_{1 \ldots N}, \tilde X_{1 \ldots N}, A_{1 \ldots N}) \triangleq \frac{1}{N_{A_j}} \sum_{i=1}^N (\tilde X_i - X_i)^{\ind{A_i = j}},
\end{align}
with $N_{A_j} \triangleq \sum_{i'} \ind{A_{i'} = j}$ is the size of the $A_j = 1$ group.

\section{Symbol Legends}\label{sec:symbol-legends}
\begin{table}[ht!]
\centering
  \begin{adjustbox}{max width=.5\textwidth}
    \begin{tabular}{r l}
      \textbf{Symbol} & \textbf{Meaning} \\
      $N$ & Number of individuals \\
      $|\mathcal{A}|$ & Number of demographic groups \\
      $A_i$ & Sensitive attribute for individual $i$ \\
      $U_{A_i}$ & Exogenous noise on sensitive attribute for individual $i$ \\
      $X_i$ & Score for individual $i$ \\
      $U_{X_i}$ & Exogenous noise on score for individual $i$ \\
      $Y_i$ & Potential outcome (loan repayment/default) for individual $i$ \\
      $U_{Y_i}$ & Exogenous noise on potential outcome for individual $i$ \\
      $T_i$ & Treatment (institution gives/withholds loan) for individual $i$ \\
      $U_{T_i}$ & Exogenous noise on treatment for individual $i$ \\
      $u_i$ & Utility of individual $i$ (from the institution's perspective) \\
      $\Delta_i$ & Expected improvement of score for individual $i$ \\
      $\tilde X_i$ & Score for individual $i$ after one time step \\
      $\mathcal{U}$ & Global utility (from institution's perspective) \\
      $\Delta_j$ & Expected change in score for group $j$ \\
    \end{tabular}
    \end{adjustbox}
  \caption{Symbol legend for Figure~\ref{fig:scm-liu}}\label{tab:scm-liu}
\end{table}

\begin{table}[hb!]
  \begin{center}
    \begin{adjustbox}{max width=.5\textwidth}
    \begin{tabular}{r l}
      \textbf{Symbol} & \textbf{Meaning} \\
      $k$ & indexes groups \\
      $P_k$ & distribution over $(X,Y)$ for group $k$ \\
      $b_k$ & expected group-$k$ baseline population growth at each step \\
      $\lambda_k^t$ & expected population for group $k$ at time $t$ \\
      $\alpha_k^t$ & mixing coeff for group $k$ at time $t$ \\
      $N^t$ & Total population at time $t$ \\
      $Z_{k}^t$ & indicator of individual belonging to $k$-th group \\
      $X^t$ & input features for an individual at time $t$ \\
      $Y^t$ & label for an individual at time $t$ \\
      $U_\theta^t$ & Exogenous noise in learning algo. (e.g., random seed) \\
      $\theta^t$ & Estimated classifier parameters at time $t$ \\
      $\hat Y^t$ & Predicted label for an individual at time $t$ \\
      $R_k^t $ & Classification error for group $k$ at time $t$ (unobserved) \\
    \end{tabular}
    \end{adjustbox}
  \end{center}
  \caption{Symbol legend for Figure~\ref{fig:scm-hashimoto}}\label{tab:scm-hashimoto}
\end{table}

Here we provide the following symbol decoders for SCMs expressed in the main paper:
\begin{itemize}
  \item Table~\ref{tab:scm-liu} decodes the symbols used in
    Figure~\ref{fig:scm-liu}
  \item Table~\ref{tab:scm-hashimoto} decodes the symbols used in
    Figure~\ref{fig:scm-hashimoto}
\end{itemize}

\section{Other SCMS}\label{sec:other-scms}

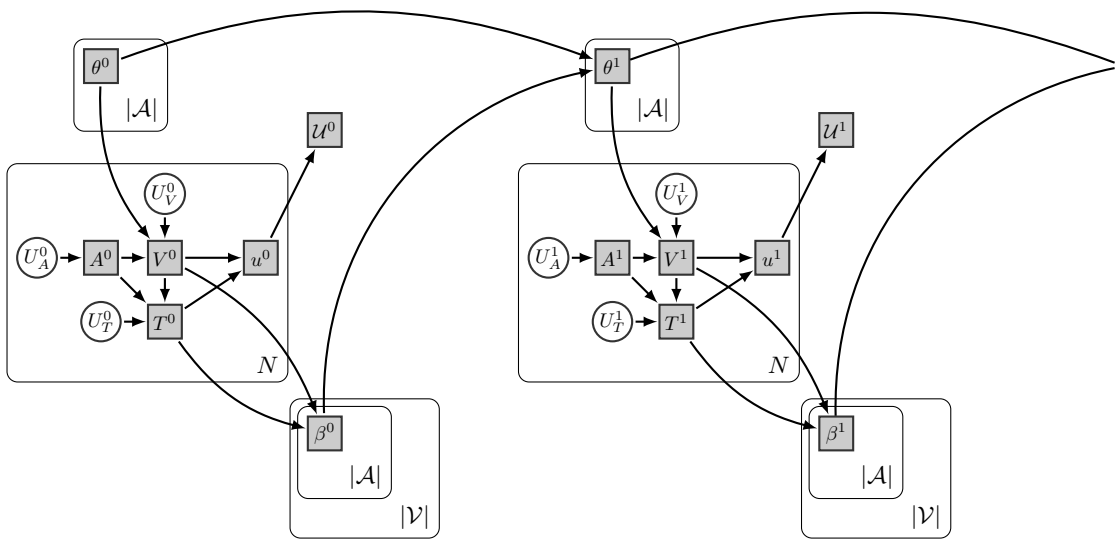
\begin{figure}[ht!]
\centering
\newcommand{\stepsize}{8}
\begin{tikzpicture}[scale=0.85, >=stealth]
\tikzstyle{empty}=[]
\tikzstyle{lat}=[circle, inner sep=1pt, minimum size = 6.5mm, thick, draw =black!80, node distance = 20mm, scale=0.75]
\tikzstyle{obs}=[circle, fill =black!20, inner sep=1pt, minimum size = 6.5mm, thick, draw =black!80, node distance = 20mm, scale=0.75]
\tikzstyle{det}=[rectangle, fill =black, inner sep=1pt, text=white, minimum size = 6mm, draw=none, node distance = 20mm, scale=0.75]
\tikzstyle{connect}=[-latex, thick]
\tikzstyle{undir}=[thick]
\tikzstyle{directed}=[->, thick, shorten >=0.5 pt, shorten <=1 pt]
\tikzstyle{lat-exo}=[circle, inner sep=1pt, minimum size = 6.5mm, thick, draw=black!80, node distance = 20mm, scale=0.75]
\tikzstyle{obs-exo}=[circle, fill=black!20, inner sep=1pt, minimum size = 6.5mm, thick, draw=black!80, node distance = 20mm, scale=0.75]
\tikzstyle{lat-box}=[rectangle, inner sep=1pt, minimum size = 6mm, draw=black!80, thick, node distance = 20mm, scale=0.75]
\tikzstyle{obs-box}=[rectangle, fill=black!20, inner sep=1pt, minimum size = 6mm, thick, draw=black!80, node distance = 20mm, scale=0.75]
ikzstyle{inv-box}=[rectangle, fill=black!20, inner sep=1pt, minimum size = 6mm, thick, draw=black!80,densely dotted, node distance = 20mm, scale=0.75]
\tikzstyle{directed}=[-latex, thick, shorten >=0.5 pt, shorten <=1 pt]
\tikzstyle{interven}=[directed, densely dotted]

\node[lat-exo] at      (0,3)    (ua0) {$U^0_{A}$};
\node[obs-box] at      (1,3)     (a0) {$A^0$};
\node[obs-box] at      (1,6)    (tj0) {$\theta^0$};
\node[obs-box] at      (2,3)     (v0) {$V^0$};
\node[lat-exo] at      (2,4)    (uv0) {$U^0_{V}$};
\node[obs-box] at      (2,2)     (t0) {$T^0$};
\node[lat-exo] at      (1,2)    (ut0) {$U^0_{T}$};
\node[obs-box] at    (4.5,5)     (U0) {$\mathcal{U}^0$};
\node[obs-box] at    (3.5,3)     (u0) {$u^0$};
\node[obs-box] at  (4.5,.25)   (bjv0) {$\beta^0$};
\node[empty]   at (5.25,.25)     (e0) {};
\node[empty]   at    (6,-.5)    (ee0) {};
\node[empty]   at  (4.25,.5)   (eee0) {};
\node[empty]   at   (1.75,6)  (eeee0) {};
\node[lat-exo] at      (0+\stepsize,3)   (ua1) {$U^1_{A}$};
\node[obs-box] at      (1+\stepsize,3)    (a1) {$A^1$};
\node[obs-box] at      (1+\stepsize,6)   (tj1) {$\theta^1$};
\node[obs-box] at      (2+\stepsize,3)    (v1) {$V^1$};
\node[lat-exo] at      (2+\stepsize,4)   (uv1) {$U^1_{V}$};
\node[obs-box] at      (2+\stepsize,2)    (t1) {$T^1$};
\node[lat-exo] at      (1+\stepsize,2)   (ut1) {$U^1_{T}$};
\node[obs-box] at    (4.5+\stepsize,5)    (U1) {$\mathcal{U}^1$};
\node[obs-box] at    (3.5+\stepsize,3)    (u1) {$u^1$};
\node[obs-box] at  (4.5+\stepsize,.25)  (bjv1) {$\beta^1$};
\node[empty]   at (5.25+\stepsize,.25)    (e1) {};
\node[empty]   at    (6+\stepsize,-.5)   (ee1) {};
\node[empty]   at  (4.25+\stepsize,.5)  (eee1) {};
\node[empty]   at   (1.75+\stepsize,6) (eeee1) {};
\node[empty] at      (1+2*\stepsize,6)  (tj2) {};
\path
        (ua0) edge [directed]  (a0)
         (a0)  edge [directed] (v0)
        (tj0)  edge [directed, bend right=20] (v0)
        (uv0) edge [directed]  (v0)
         (v0)  edge [directed] (t0)
         (a0)  edge [directed] (t0)
        (ut0) edge [directed]  (t0)
         (t0)  edge [directed] (u0)
         (v0)  edge [directed] (u0)
         (u0)  edge [directed] (U0)
         (t0)  edge [directed, bend right=20] (bjv0)
         (v0)  edge [directed, bend left=20] (bjv0)
        (tj0)   edge [directed, bend left=20] (tj1)
        (bjv0)  edge [directed, bend left=40] (tj1)
        (ua1) edge [directed]  (a1)
         (a1)  edge [directed] (v1)
        (tj1)  edge [directed, bend right=20] (v1)
        (uv1) edge [directed]  (v1)
         (v1)  edge [directed] (t1)
         (a1)  edge [directed] (t1)
        (ut1) edge [directed]  (t1)
         (t1)  edge [directed] (u1)
         (v1)  edge [directed] (u1)
         (u1)  edge [directed] (U1)
         (t1)  edge [directed, bend right=20] (bjv1)
         (v1)  edge [directed, bend left=20] (bjv1)
        (tj1)   edge [thick, bend left=20] (tj2)
        (bjv1)  edge [thick, bend left=40] (tj2)
        ;
\plate {} {(ua0)(a0)(v0)(uv0)(t0)(ut0)(u0)} {$N$} ;
\plate {} {(tj0)(eeee0)} {$|\mathcal{A}|$} ;
\plate {} {(bjv0)(e0)} {$|\mathcal{A}|$} ;
\plate {} {(bjv0)(ee0)(eee0)} {$|\mathcal{V}|$} ;
\plate {} {(ua1)(a1)(v1)(uv1)(t1)(ut1)(u1)} {$N$} ;
\plate {} {(tj1)(eeee1)} {$|\mathcal{A}|$} ;
\plate {} {(bjv1)(e1)} {$|\mathcal{A}|$} ;
\plate {} {(bjv1)(ee1)(eee1)} {$|\mathcal{V}|$} ;
\end{tikzpicture}
\caption{
    SCM for the group dynamics model proposed by \citet{mouzannar2019fair}.
    See Table~\ref{tab:scm-mouzannar} for a description of each symbol.
  }\label{fig:scm-mouzannar}
\end{figure}

\begin{figure}[ht!]
\centering
\noindent\resizebox{.75\textwidth}{!}{
\newcommand{\stepsize}{8}
\begin{tikzpicture}[scale=0.85, >=stealth]
\tikzstyle{empty}=[]
\tikzstyle{lat}=[circle, inner sep=1pt, minimum size = 6.5mm, thick, draw =black!80, node distance = 20mm, scale=0.75]
\tikzstyle{obs}=[circle, fill=black!20, inner sep=1pt, minimum size = 6.5mm, thick, draw=black!80, node distance = 20mm, scale=0.75]
\tikzstyle{det}=[rectangle, fill=black!20, inner sep=1pt, minimum size = 6mm, thick, draw=black!80, node distance = 20mm, scale=0.75, font=\huge]
\tikzstyle{connect}=[-latex, thick]
\tikzstyle{undir}=[thick]
\tikzstyle{directed}=[->, thick, shorten >=0.5 pt, shorten <=1 pt]

\node[det] at (0,-1.5)  (ui0) {$u^0$};
\node[det] at (1,4)  (aj0) {$a^0$};
\node[det] at (1,2)  (dij0) {$d^0$};
\node[det] at (1,0)  (rij0) {$r^0$};
\node[obs] at (-1,2)  (urec0) {$U_{\text{reco}}^0$};
\node[lat] at (2,5)  (k0) {$k$};
\node[lat] at (3,5)  (d0) {$\delta$};
\node[lat] at (4,5)  (b0) {$\beta$};
\node[det] at (3,2)  (vij0) {$v^0$};
\node[det] at (3,-1.5)  (wi0) {$w^0$};
\node[lat] at (2,-3.25)  (l0) {$\lambda$};
\node[lat] at (3,-3.25)  (tp0) {$\theta'$};
\node[lat] at (4,-3.25)  (t0) {$\theta$};
\node[det] at (4,1)  (cij0) {$c^0$};
\node[det] at (5,4)  (zj0) {$z^0$};
\node[lat] at (2.5,1)  (uchoiceij0) {$U_{\text{choice}}^0$};
\node[lat] at  (0+\stepsize-1.5,-1.5)  (udrifti1) {$U_{\text{drift}}^1$};
\node[lat] at  (5+\stepsize+1,4)  (p1) {$p$};
\node[det] at  (0+\stepsize,-1.5)  (ui1) {$u^1$};
\node[lat] at (0+\stepsize+1,-1.5)  (ts1) {$\theta^*$};
\node[det] at  (1+\stepsize,4)  (aj1) {$a^1$};
\node[det] at  (1+\stepsize,2)  (dij1) {$d^1$};
\node[det] at  (1+\stepsize,0)  (rij1) {$r^1$};
\node[obs] at (-1+\stepsize,2)  (urec1) {$U_{\text{reco}}^1$};
\node[lat] at  (2+\stepsize,5)  (k1) {$k$};
\node[lat] at  (3+\stepsize,5)  (d1) {$\delta$};
\node[lat] at  (4+\stepsize,5)  (b1) {$\beta$};
\node[det] at  (3+\stepsize,2)  (vij1) {$v^1$};
\node[det] at  (3+\stepsize,-1.5)  (wi1) {$w^1$};
\node[lat] at  (2+\stepsize,-3.25)  (l1) {$\lambda$};
\node[lat] at  (3+\stepsize,-3.25)  (tp1) {$\theta'$};
\node[lat] at  (4+\stepsize,-3.25)  (t1) {$\theta$};
\node[det] at  (4+\stepsize,1)  (cij1) {$c^1$};
\node[det] at  (5+\stepsize,4)  (zj1) {$z^1$};
\node[lat] at  (2.5+\stepsize,1)  (uchoiceij1) {$U_{\text{choice}}^1$};
\node[lat] at  (0+2*\stepsize-1.5,-1.5)  (udrifti2) {$U_{\text{drift}}^2$};
\node[lat] at  (5+2*\stepsize+1,4)  (p2) {$p$};
\node[det] at  (0+2*\stepsize,-1.5)  (ui2) {$u^2$};
\node[lat] at  (0+2*\stepsize+1,-1.5) (ts2) {$\theta^*$};
\node[det] at  (1+2*\stepsize,4)  (aj2) {$a^2$};
\node[det] at  (1+2*\stepsize,2)  (dij2) {$d^2$};
\node[det] at  (1+2*\stepsize,0)  (rij2) {$r^2$};
\node[obs] at (-1+2*\stepsize,2)  (urec2) {$U_{\text{reco}}^2$};
\node[lat] at  (2+2*\stepsize,5)  (k2) {$k$};
\node[lat] at  (3+2*\stepsize,5)  (d2) {$\delta$};
\node[lat] at  (4+2*\stepsize,5)  (b2) {$\beta$};
\node[det] at  (3+2*\stepsize,2)  (vij2) {$v^2$};
\node[det] at  (3+2*\stepsize,-1.5)  (wi2) {$w2$};
\node[lat] at  (2+2*\stepsize,-3.25)  (l2) {$\lambda$};
\node[lat] at  (3+2*\stepsize,-3.25)  (tp2) {$\theta'$};
\node[lat] at  (4+2*\stepsize,-3.25)  (t2) {$\theta$};
\node[det] at  (4+2*\stepsize,1)  (cij2) {$c^2$};
\node[det] at  (5+2*\stepsize,4)  (zj2) {$z^2$};
\node[lat] at  (2.5+2*\stepsize,1)  (uchoiceij2) {$U_{\text{choice}}^2$};
\path
        (ui0) edge [directed, bend left=20] (dij0)
        (aj0) edge [directed] (dij0)
        (ui0) edge [directed] (rij0)
        (aj0) edge [directed, bend right=20] (rij0)
        (urec0) edge [directed] (rij0)
        (rij0) edge [directed, bend right=20] (wi0)
        (t0)   edge [directed] (wi0)
        (tp0)  edge [directed] (wi0)
        (l0)   edge [directed] (wi0)
        (zj0)  edge [directed, bend left=25] (wi0)
        (dij0) edge [directed, bend right=20] (wi0)
        (dij0) edge [directed] (vij0)
        (k0)   edge [directed] (vij0)
        (d0)   edge [directed] (vij0)
        (b0)   edge [directed] (vij0)
        (vij0) edge [directed] (cij0)
        (wi0)  edge [directed] (cij0)
        (uchoiceij0)  edge [directed] (cij0)
        (ui0)  edge [directed, bend right=90] (ui1)
        (cij0) edge [directed, bend left=5] (ui1)
        (dij0) edge [directed, bend left=60] (ui1)
        (ts1)  edge [directed              ] (ui1)
        (zj0) edge  [directed, bend left=90] (zj1)
        (udrifti1) edge [directed] (ui1)
        (ui1) edge [directed, bend left=20] (dij1)
        (aj1) edge [directed] (dij1)
        (ui1) edge [directed] (rij1)
        (aj1) edge [directed, bend right=20] (rij1)
        (urec1) edge [directed] (rij1)
        (rij1) edge [directed, bend right=20] (wi1)
        (t1)   edge [directed] (wi1)
        (tp1)  edge [directed] (wi1)
        (l1)   edge [directed] (wi1)
        (zj1)  edge [directed, bend left=25] (wi1)
        (p1)   edge [directed] (zj1)
        (dij1) edge [directed, bend right=20] (wi1)
        (dij1) edge [directed] (vij1)
        (k1)   edge [directed] (vij1)
        (d1)   edge [directed] (vij1)
        (b1)   edge [directed] (vij1)
        (vij1) edge [directed] (cij1)
        (wi1)  edge [directed] (cij1)
        (uchoiceij1)  edge [directed] (cij1)
        (ui1)  edge [directed, bend right=90] (ui2)
        (cij1) edge [directed, bend left=5] (ui2)
        (dij1) edge [directed, bend left=60] (ui2)
        (ts2)  edge [directed              ] (ui2)
        (zj1) edge  [directed, bend left=90] (zj2)
        (udrifti2) edge [directed] (ui2)
        (ui2) edge [directed, bend left=20] (dij2)
        (aj2) edge [directed] (dij2)
        (ui2) edge [directed] (rij2)
        (aj2) edge [directed, bend right=20] (rij2)
        (urec2) edge [directed] (rij2)
        (rij2) edge [directed, bend right=20] (wi2)
        (t2)   edge [directed] (wi2)
        (tp2)  edge [directed] (wi2)
        (l2)   edge [directed] (wi2)
        (zj2)  edge [directed, bend left=25] (wi2)
        (p2)   edge [directed] (zj2)
        (dij2) edge [directed, bend right=20] (wi2)
        (dij2) edge [directed] (vij2)
        (k2)   edge [directed] (vij2)
        (d2)   edge [directed] (vij2)
        (b2)   edge [directed] (vij2)
        (vij2) edge [directed] (cij2)
        (wi2)  edge [directed] (cij2)
        (uchoiceij2)  edge [directed] (cij2)
        ;
\plate {} {(rij0)(ui0)(wi0)(dij0)(cij0)(uchoiceij0)} {$|\mathcal{U}|$} ; %
\plate {} {(rij0)(aj0)(dij0)(cij0)(uchoiceij0)(zj0)} {$|\mathcal{A}|$} ; %
\plate {} {(rij1)(ui1)(wi1)(dij1)(cij1)(uchoiceij1)} {$|\mathcal{U}|$} ; %
\plate {} {(rij1)(aj1)(dij1)(cij1)(uchoiceij1)(zj1)} {$|\mathcal{A}|$} ; %
\plate {} {(rij2)(ui2)(wi2)(dij2)(cij2)(uchoiceij2)} {$|\mathcal{U}|$} ; %
\plate {} {(rij2)(aj2)(dij2)(cij2)(uchoiceij2)(zj2)} {$|\mathcal{A}|$} ; %
\end{tikzpicture}
}
\caption{
    SCM for the news recommendation simulator model proposed by
     \citet{bountouridis2019siren}.
    The key dynamic modeling is in the user vectors in topic space, which drift
     over time towards the articles that are consumed (these in turn partially
     depend on the recommendations).
    Articles are also modeled as decaying in popularity in time.
    See Table~\ref{tab:scm-bountouridis} for explanation of all symbols.
 }\label{fig:scm-bountouridis}
\end{figure}
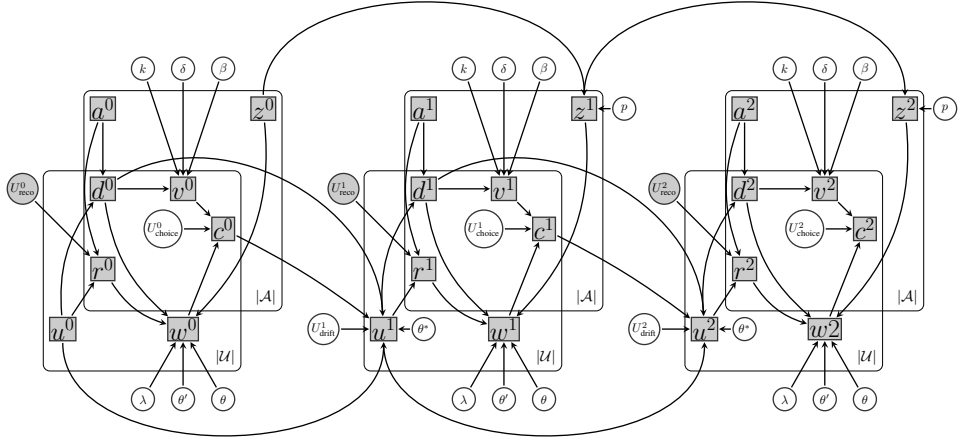

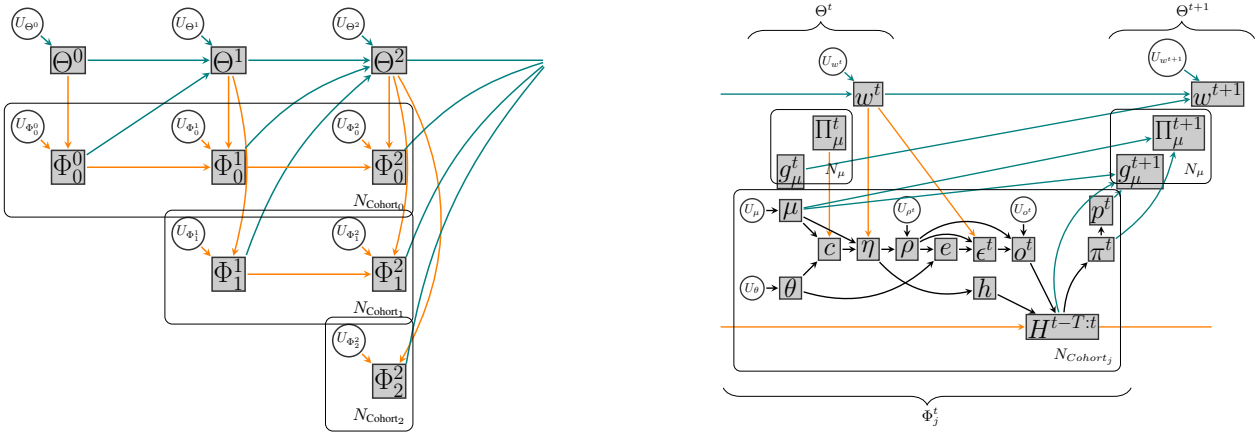
\begin{figure*}[t!]
  \centering
  \begin{subfigure}[t]{0.45\textwidth}
    \noindent\resizebox{\textwidth}{!}{
      \newcommand{\ssi}{3}  %
\newcommand{\csi}{2}  %
\newcommand{\exu}{.3cm}  %
\begin{tikzpicture}[scale=1., >=stealth]
\tikzstyle{empty}=[]
\tikzstyle{lat-exo}=[circle, inner sep=1pt, minimum size = 6.5mm, thick, draw=black!80, node distance = 20mm, scale=0.75]
\tikzstyle{obs-exo}=[circle, fill=black!20, inner sep=1pt, minimum size = 6.5mm, thick, draw=black!80, node distance = 20mm, scale=0.75]
\tikzstyle{lat-box}=[rectangle, inner sep=1pt, minimum size = 6mm, draw=black!80, thick, node distance = 20mm, scale=0.75]
\tikzstyle{obs-box}=[rectangle, fill=black!20, inner sep=1pt, minimum size = 6mm, thick, draw=black!80, node distance = 20mm, scale=0.75, font=\huge]
\tikzstyle{inv-box}=[rectangle, fill=black!20, inner sep=1pt, minimum size = 6mm, thick, draw=black!80,densely dotted, node distance = 20mm, scale=0.75]
\tikzstyle{connect}=[-latex, thick]
\tikzstyle{undir}=[thick]
\tikzstyle{directed}=[->, thick, shorten >=0.5 pt, shorten <=1 pt]
\tikzstyle{undir-cohort}=[thick, orange]
\tikzstyle{directed-cohort}=[->, thick, shorten >=0.5 pt, shorten <=1 pt, orange]
\tikzstyle{undir-global}=[thick, teal]
\tikzstyle{directed-global}=[->, thick, shorten >=0.5 pt, shorten <=1 pt, teal]
\tikzstyle{undir}=[thick]
\tikzstyle{directed}=[->, thick, shorten >=0.5 pt, shorten <=1 pt]
\tikzstyle{semi-box}=[rectangle with diagonal fill, diagonal top color=black!20, diagonal bottom color=white, diagonal from left to right, draw]
\tikzstyle{interven}=[directed, densely dotted]

\node[obs-box]   (t0)                            at (0,0) {$\Theta^0$};
\node[lat-exo]  (ut0)         [above left=\exu of t0] {$U_{\Theta^0}$};
\node[obs-box]  (p00)                    at (0,-\csi) {$\Phi^0_{0}$};
\node[lat-exo] (up00)    [above left=\exu of p00] {$U_{\Phi^0_{0}}$};
\node[obs-box]   (t1)                            at (\ssi,0) {$\Theta^1$};
\node[lat-exo]  (ut1)            [above left=\exu of t1] {$U_{\Theta^1}$};
\node[obs-box]  (p10)                    at (\ssi,-\csi) {$\Phi^1_{0}$};
\node[lat-exo] (up10)       [above left=\exu of p10] {$U_{\Phi^1_{0}}$};
\node[obs-box]  (p11)                  at (\ssi,-2*\csi) {$\Phi^1_{1}$};
\node[lat-exo] (up11)       [above left=\exu of p11] {$U_{\Phi^1_{1}}$};
\node[obs-box]   (t2)                            at (2*\ssi,0) {$\Theta^2$};
\node[lat-exo]  (ut2)              [above left=\exu of t2] {$U_{\Theta^2}$};
\node[obs-box]  (p20)                    at (2*\ssi,-\csi) {$\Phi^2_{0}$};
\node[lat-exo] (up20)         [above left=\exu of p20] {$U_{\Phi^2_{0}}$};
\node[obs-box]  (p21)                  at (2*\ssi,-2*\csi) {$\Phi^2_{1}$};
\node[lat-exo] (up21)         [above left=\exu of p21] {$U_{\Phi^2_{1}}$};
\node[obs-box]  (p22)                  at (2*\ssi,-3*\csi) {$\Phi^2_{2}$};
\node[lat-exo] (up22)         [above left=\exu of p22] {$U_{\Phi^2_{2}}$};
\node[empty]    (t3)                            at (3*\ssi,0)            {};
  \path
     (ut0)    edge [directed-global]                              (t0)
     (t0)     edge [directed-cohort]                      (p00)
     (up00)   edge [directed-cohort]                      (p00)
     (t0)     edge [directed-global]                       (t1)
     (p00)    edge [directed-global]                       (t1)
     (p00)    edge [directed-cohort]                      (p10)
     (ut1)    edge [directed-global]                              (t1)
     (up10)   edge [directed-cohort]                      (p10)
     (up11)   edge [directed-cohort]                      (p11)
     (t1)     edge [directed-cohort]                      (p10)
     (t1)     edge [directed-cohort, bend left=15]        (p11)
     (t1)     edge [directed-global]                       (t2)
     (p10)    edge [directed-global, bend left=15]         (t2)
     (p11.north east) edge [directed-global, bend left=15] (t2)
     (p10)    edge [directed-cohort]                      (p20)
     (p11)    edge [directed-cohort]                      (p21)
     (ut2)    edge [directed-global]                       (t2)
     (up20)   edge [directed-cohort]                      (p20)
     (up21)   edge [directed-cohort]                      (p21)
     (up22)   edge [directed-cohort]                      (p22)
     (t2)     edge [directed-cohort]                      (p20)
     (t2)     edge [directed-cohort, bend left=15]        (p21)
     (t2)     edge [directed-cohort, bend left=30]        (p22)
     (t2)        edge [undir-global]                       (t3)
     (p20)       edge [undir-global, bend left=15]         (t3)
     (p21.north east)    edge [undir-global, bend left=15] (t3)
     (p22.north east)    edge [undir-global, bend left=15] (t3)
        ;
\plate {cohort0} {(p00)(up00)(p10)(up10)(p20)(up20)} {$N_{\text{Cohort}_0}$} ;
\plate {cohort1} {(p11)(up11)(p21)(up21)}            {$N_{\text{Cohort}_1}$} ;
\plate {cohort2} {(p22)(up22)}                       {$N_{\text{Cohort}_2}$} ;

\end{tikzpicture}
    }
  \caption{
  Macro-level DAG showing how market state $\Theta^t$ and worker
    cohorts $\Phi_{j}^t$ dynamically affect one another}\label{fig:scm-hu-macro}
  \end{subfigure}%
  \hfill
  \begin{subfigure}[t]{0.45\textwidth}
    \noindent\resizebox{\textwidth}{!}{
      \newcommand{\ssi}{9}  %
\newcommand{\exu}{.3cm}  %
\begin{tikzpicture}[scale=.8, >=stealth]
\tikzstyle{empty}=[]
\tikzstyle{lat-exo}=[circle, inner sep=1pt, minimum size = 6.5mm, thick, draw=black!80, node distance = 20mm, scale=0.75]
\tikzstyle{obs-exo}=[circle, fill=black!20, inner sep=1pt, minimum size = 6.5mm, thick, draw=black!80, node distance = 20mm, scale=0.75]
\tikzstyle{lat-box}=[rectangle, inner sep=1pt, minimum size = 6mm, draw=black!80, thick, node distance = 20mm, scale=0.75]
\tikzstyle{obs-box}=[rectangle, fill=black!20, inner sep=1pt, minimum size = 6mm, thick, draw=black!80, node distance = 20mm, scale=0.75, font=\huge]
\tikzstyle{inv-box}=[rectangle, fill=black!20, inner sep=1pt, minimum size = 6mm, thick, draw=black!80,densely dotted, node distance = 20mm, scale=0.75]
\tikzstyle{connect}=[-latex, thick]
\tikzstyle{undir}=[thick]
\tikzstyle{directed}=[->, thick, shorten >=0.5 pt, shorten <=1 pt]
\tikzstyle{undir-cohort}=[thick, orange]
\tikzstyle{directed-cohort}=[->, thick, shorten >=0.5 pt, shorten <=1 pt, orange]
\tikzstyle{undir-global}=[thick, teal]
\tikzstyle{directed-global}=[->, thick, shorten >=0.5 pt, shorten <=1 pt, teal]
\tikzstyle{semi-box}=[rectangle with diagonal fill, diagonal top color=black!20, diagonal bottom color=white, diagonal from left to right, draw]
\tikzstyle{interven}=[directed, densely dotted]

\node[]  (Hin)                                              at (-2,-2)  {};
\node[]  (wn)                                                at (-2,4)  {};
\node[obs-box]  (Pm0)                               at (1,3) {$\Pi_\mu^t$};
\node[obs-box]   (w0)                                     at (2,4) {$w^t$};
\node[lat-exo]  (uw0)                  [above left=\exu of w0] {$U_{w^t}$};
\node[obs-box]  (gm0)                                 at (0,2) {$g_\mu^t$};
\node[obs-box]  (mi0)                                     at (0,1) {$\mu$};
\node[lat-exo] (umi0)                       [left=\exu of mi0] {$U_{\mu}$};
\node[obs-box]  (ti0)                                 at (0,-1) {$\theta$};
\node[lat-exo] (uti0)                    [left=\exu of ti0] {$U_{\theta}$};
\node[obs-box]  (ci0)                                       at (1,0) {$c$};
\node[obs-box]  (ni0)                                    at (2,0) {$\eta$};
\node[obs-box]  (ri0)                                    at (3,0) {$\rho$};
\node[obs-box]  (ei0)                                       at (4,0) {$e$};
\node[obs-box] (epi0)                              at (5,0) {$\epsilon^t$};
\node[obs-box] (hi0)                                       at (5,-1) {$h$};
\node[lat-exo] (uri0)                   [above=\exu of ri0] {$U_{\rho^t}$};
\node[obs-box]  (oi0)                                     at (6,0) {$o^t$};
\node[lat-exo] (uoi0)                      [above=\exu of oi0] {$U_{o^t}$};
\node[obs-box]  (Hi0)                             at (7, -2) {$H^{t-T:t}$};
\node[obs-box]  (Pi0)                                   at (8,0) {$\pi^t$};
\node[obs-box]  (pi0)                                     at (8,1) {$p^t$};
\node[obs-box]  (Pm1)                     at (\ssi+1,3)  {$\Pi_\mu^{t+1}$};
\node[obs-box]   (w1)                            at (\ssi+2,4) {$w^{t+1}$};
\node[lat-exo]  (uw1)              [above left=\exu of w1] {$U_{w^{t+1}}$};
\node[obs-box]  (gm1)                         at (\ssi,2)  {$g_\mu^{t+1}$};
\node[]  (Hi1)                                          at (\ssi+2,-2)  {};

  \path
         (Hin)                edge [directed-cohort] (Hi0)
          (wn)                 edge [directed-global] (w0)
         (uw0)                 edge [directed-global] (w0)
         (Pm0)              edge [directed-cohort] (ci0)
          (w0)              edge [directed-cohort] (ni0)
          (w0)             edge [directed-cohort] (epi0)
        (umi0)                       edge [directed] (mi0)
         (mi0)                       edge [directed] (ci0)
         (mi0)                       edge [directed] (ni0)
        (uti0)                       edge [directed] (ti0)
         (ti0)                       edge [directed] (ci0)
         (ci0)                       edge [directed] (ni0)
         (ti0)        edge [directed, bend right=30] (ei0)
         (ni0)                       edge [directed] (ri0)
         (ni0)        edge [directed, bend right=30] (hi0)
         (hi0)                       edge [directed] (Hi0)
         (ri0)                       edge [directed] (ei0)
        (uri0)                       edge [directed] (ri0)
         (ri0)        edge [directed, bend left=30] (epi0)
         (ri0)         edge [directed, bend left=40] (oi0)
         (ei0)                      edge [directed] (epi0)
         (epi0)                      edge [directed] (oi0)
        (uoi0)                       edge [directed] (oi0)
         (oi0)                       edge [directed] (Hi0)
         (Hi0)         edge [directed, bend left=20] (Pi0)
         (Pi0)                       edge [directed] (pi0)
          (w0)                 edge [directed-global] (w1)
         (gm0)                 edge [directed-global] (w1)
         (mi0)                edge [directed-global] (gm1)
         (mi0)                edge [directed-global] (Pm1)
         (Hi0)  edge [directed-global, bend left=40] (gm1)
         (pi0)                edge [directed-global] (gm1)
         (Pi0) edge [directed-global, bend right=20] (Pm1)
         (Hi0)                   edge [undir-cohort] (Hi1)
         (uw1)                 edge [directed-global] (w1)
        ;
\plate {Theta0} {(Pm0)(gm0.north west)} {$N_{\mu}$} ;
\plate {Theta1} {(Pm1)(gm1.north west)} {$N_{\mu}$} ;
\plate {Cohort} {(ci0) (mi0) (umi0) (ti0) (uti0) (ci0) (ni0) (ri0) (ei0) (epi0) (oi0) (Hi0) (Pi0) (pi0) } {$N_{Cohort_j}$} ;
\node[empty]   (ew0)              [above right=\exu of w0] {};
\node[empty]   (eew0)                  [above=\exu of ew0] {};
\node[empty]   (egm0)                    [left=3cm of eew0] {};
\draw [decorate,decoration={brace,amplitude=10pt},xshift=0pt,yshift=0pt] (egm0) -- (eew0) node [text width=1cm,black,midway,xshift=0.4cm,yshift=0.6cm] {\footnotesize $\Theta^{t}$};
\node[empty]   (ew1)              [above right=\exu of w1] {};
\node[empty]   (eew1)                  [above=\exu of ew1] {};
\node[empty]   (egm1)                    [left=3cm of eew1] {};
\draw [decorate,decoration={brace,amplitude=10pt},xshift=0pt,yshift=0pt] (egm1)
-- (eew1) node [text width=1cm,black,midway,xshift=0.4cm,yshift=0.6cm] {\footnotesize $\Theta^{t+1}$};
\node[empty]   (ecl0)           [below left=\exu of Cohort] {};
\node[empty]   (ecr0)          [below right=\exu of Cohort] {};
\draw [decorate,decoration={brace,amplitude=10pt},xshift=0pt,yshift=0pt] (ecr0) -- (ecl0) node [text width=1cm,black,midway,xshift=0.4cm,yshift=-0.6cm] {\footnotesize $\Phi_{j}^t$};
\node[empty]   (ecl0)           [below left=\exu of Cohort] {};
\end{tikzpicture}
    }
  \caption{Micro-level DAG isolating how market state $\Theta^t$ affects
    investment and effort levels of a single worker cohort, and how worker
    choices affect market state at the next step}\label{fig:scm-hu-micro}
  \end{subfigure}%
  \caption{
    SCM for the hiring model from \citet{hu2018short}.
    \ref{fig:scm-hu-macro} shows macro-level causal assumptions.
    At step $t$ the global state $\Theta^{t}$ of the PLM affects the choices
    of all cohorts of workers (a cohort denotes workers that enter the market at
    the same step) via wage signals (\ref{fig:scm-hu-micro}).
    The choices of investment and effort and resulting outcomes in turn affect
    the workers themselves in terms of hiring decisions, and the global state of
    the market in terms of average group reputation and performance per group.
    {\color{teal}Teal} arrows denote structural functions going into the global state.
    {\color{orange}Orange} arrows denote structural functions going into the cohort state.
    {\color{black}Black} arrows denote structural functions within the cohort state.
    See Table~\ref{tab:scm-hu} in Appendix~\ref{sec:symbol-legends} for
      explanation of all symbols.
      the dynamics.
  }\label{fig:scm-hu}
\end{figure*}

Here we provide some SCMs for some additional papers from the literature:
\begin{itemize}
  \item Figure~\ref{fig:scm-mouzannar} describes the multi-step loan setting
    discussed by \citet{mouzannar2019fair}.
    Their model is similar to the one proposed by \citet{liu2018delayed}.
    The main difference is that \citet{mouzannar2019fair} describes dynamics
    that unfold \emph{exclusively} at the population level, where decisions
    rendered by the institution do not affect the future well-being of the
    individuals themselves.
  \item Figure~\ref{fig:scm-bountouridis} corresponds to the news recommender
    simulator discussed by \citet{bountouridis2019siren}.
    The goal of this simulator was to understand the long-term effects of
    recommender algorithms on news consumption behaviors.
  \item Figure~\ref{fig:scm-hu} shows the hiring market model proposed in \citet{hu2018short}.
    Figure~\ref{fig:scm-hu-macro} shows the higher-level structure of the model: a
    global state of the hiring market $\Theta$ progresses through time, a cohort of
    workers are initialized at each time step with attributes $\Phi$ set by the
    current global state, and the cohorts progress through time, feeding back into
    the global state at each step.
    
    Figure~\ref{fig:scm-hu-micro} shows the structure of each individual/cohort's
    journey through the labour market.
    At the top of Figure~\ref{fig:scm-hu-micro}, we see the variables which
    constitute the global state $\Theta$: wages $w$, reputation $\Pi_{\mu}$ of group
    $\mu$, and the proportion of ``good'' workers on the permanent labour market in
    group $\mu$,  $g_{\mu}$.
    The bottom plate of Figure~\ref{fig:scm-hu-micro} shows the variables which are
    part of $\Phi$ and which correspond to attributes of an individual worker's
    experience.
\end{itemize}

\subsection{Symbols for Figures in Supplemental Material}
\begin{table}[hb!]
\centering
  \begin{adjustbox}{max width=.5\textwidth}
    \begin{tabular}{r l}
      \textbf{Symbol} & \textbf{Meaning} \\
      $t$ & indexes time \\
      $i$ & indexes individuals \\
      $j$ & indexes cohorts \\
      $w^t$ & wages at time $t$\\
      $g^t_\mu$ & proportion ``good'' group-$\mu$ workers in PLM\\
      $\Pi_\mu^t$ & group $\mu$ reputation at time $t$ \\
      $\mu_i$ & group membership for worker $i$ \\
      $\theta_i$ & individual $i$ ability\\
      $c_i$ & cost of investment for individual $i$\\
      $\eta_i$ & investment level for individual $i$\\
      $\rho_i$ & qualification level for individual $i$\\
      $e_i$ & individual-$i$ cost of effort \\
      $\epsilon_i^t$ & individual-$i$ actual effort exerted at time $t$\\
      $o_i^t$ & individual-$i$ outcome at time $t$\\
      $h_i$ & was individual hired to TLM following education?\\
      $H_i^{t-\tau:t-1}$ & individual-$i$ $\tau$-recent history (outcomes and TLM/PLM status)\\
      $\pi_i^t$ & individual $i$ reputation at time $t$ \\
      $p_i^t$ & was individual hired to PLM at step $t$?\\
    \end{tabular}
    \end{adjustbox}
  \caption{Symbol legend for Figure~\ref{fig:scm-hu}}\label{tab:scm-hu}
\end{table}

\begin{table}[hb!]
\centering
\begin{adjustbox}{max width=.5\textwidth}
    \begin{tabular}{r l}
      \textbf{Symbol} & \textbf{Meaning} \\
      $A_i$ & Sensitive attribute for individual $i$ \\
      $U_{A_i}$ & Exogenous noise on sensitive attribute for individual $i$ \\
      $|\mathcal{A}|$ & Number of demographic groups \\
      $V_i$ & Qualification for individual $i$ \\
      $U_{V_i}$ & Exogenous noise on qualification for individual $i$ \\
      $|\mathcal{V}|$ & Number of qualification levels \\
      $\theta_j^t$ & Bernoulli parameter of qualifications of group $j$ at time $t$ \\
      $N$ & Number of individuals \\
      $T_i$ & ``Treatment'' (whether the institution gives loan) for individual $i$ \\
      $U_{T_i}$ & Exogenous noise on treatment for individual $i$ \\
      $u_i$ & Utility of individual $i$ (from the institution's perspective) \\
      $\beta^t_{j,v}$ & Selection rate for group $j$ members with qual. $v$ at step $t$ \\
      $\mathcal{U}$ & Global institutional utility\\
    \end{tabular}
    \end{adjustbox}
  \caption{Symbol legend for Figure~\ref{fig:scm-mouzannar}}\label{tab:scm-mouzannar}
\end{table}

\begin{table}[hb!]
\centering
  \begin{adjustbox}{max width=.5\textwidth}
    \begin{tabular}{r@{}l}
      &
      \begin{tabular}{@{} r l @{}}
        \textbf{Symbol   } & \textbf{Meaning} \\
      \end{tabular} \\
      User & $\left\{
      \begin{tabular}{@{} r l @{}}
        \ \ \ \ \ \ \ \ $u_i^t$& $i$-th user topic vector at step $t$ \\
        $\theta$ & Awareness decay with user-article distance \\
        $\theta'$ & Awareness decay with article prominence \\
        $\lambda$ & Prominent vs proximity in awareness computation \\
        $w$ & Max awareness pool size for any user \\
        $k$ & $i$-th user's sensitivity to article proximity in awareness computation \\
        $\theta^*_i$ & $i$-th user's sensitivity to article proximity in drift computation \\
        $s$ & number of articles read per user per step \\
        $U_{\text{drift},i}^t$ & Exogenous noise on user $i$'s drift at step $t$\\
        $|\mathcal{U}|$ & Number of users \\
      \end{tabular} \right.\kern-\nulldelimiterspace $ \\
      Article & $\left\{
      \begin{tabular}{@{} r l @{}}
        \ \ \ \ \ \ \ \ $a_j$ & $j$-th article topic vector \\
        $z^0_j$ & initial prominence of article $j$ (possibly shared across topic) \\
        $z^t_j$ & prominence of article $j$ at step $t$ \\
        $p$ & prominence (linear) decay factor \\
        $|\mathcal{A}|$ & Number of articles \\
      \end{tabular} \right.\kern-\nulldelimiterspace $ \\
      User-article & $\left\{
      \begin{tabular}{@{} r l @{}}
        $d_{i,j}^t$ & distance between user $u_i$ and article $a_j$ at step $t$ \\
        $v_{i,j}^t$ & computed step-$t$ distance (user $u_i$, article $a_j$) in awareness computation\\
        $c_{i,j}^t$ & computed step-$t$ choice of user $u_i$ about article $a_j$ \\
        $U_{\text{choice},i,j}^t$ & Exogenous noise on user $i$'s choice of article $j$ at step $t$\\
      \end{tabular} \right.\kern-\nulldelimiterspace $ \\
      Recommender & $\left\{
      \begin{tabular}{@{} r l @{}}
        \ \ \ \ \ \ \ \ $m$ & Number of articles recommended to each user at each step \\
        $\kappa_{i,j}^t$ & Rank of recommendation of article $j$ to user $i$ at step $t$\\
        $\delta$ & Base amount of salience boost induced by a recommendation \\
        $\beta$ & Rank-decay of salience induced by a recommendation \\
        $U_{\text{reco}}^t$ & Exogenous (possibly observed) noise in the recommender algo at step $t$ \\
        $d$ & Number of steps in the simulation \\
      \end{tabular} \right.\kern-\nulldelimiterspace $ \\
    \end{tabular}
  \end{adjustbox}
  \caption{Symbol legend for Figure \ref{fig:scm-bountouridis}}
  \label{tab:scm-bountouridis}
\end{table}

Here we provide the following symbol decoders for SCMs expressed in the Appendices:
\begin{itemize}
  \item Table~\ref{tab:scm-hu} decodes the symbols used in
    Figure~\ref{fig:scm-hu}
  \item Table~\ref{tab:scm-mouzannar} decodes the symbols used in
    Figure~\ref{fig:scm-mouzannar}
  \item Table~\ref{tab:scm-bountouridis} decodes the symbols used in
    Figure~\ref{fig:scm-bountouridis}
\end{itemize}

\end{document}